\documentclass[10pt,twocolumn,letterpaper]{article}
\usepackage{cvpr}
\usepackage{times}
\usepackage{epsfig}
\usepackage{graphicx}
\usepackage{amsmath}
\usepackage{amssymb}
\usepackage[pagebackref=true,breaklinks=true,letterpaper=true,colorlinks,bookmarks=false]{hyperref}
\usepackage{enumitem}
\usepackage{multirow}
\usepackage{pifont}
\usepackage{color}
\cvprfinalcopy 
\def\cvprPaperID{1648} 

\newcommand*\samethanks[1][\value{footnote}]{\footnotemark[#1]}

\begin{document}
	
	\title{Feedback Network for Image Super-Resolution}

\author{Zhen Li$^{1}$ \quad Jinglei Yang$^{2}$ \quad Zheng Liu$^{3}$ \quad Xiaomin Yang$^{1}$\thanks{Corresponds to: {\tt\footnotesize \{arielyang,wuwei\}@scu.edu.cn}} \quad Gwanggil Jeon$^{4}$ \quad Wei Wu$^{1}$\samethanks\\
	$^1$Sichuan University, $^2$University of California, Santa Barbara, $^3$University of British Columbia,\\ $^4$Incheon National University	
}	

	\maketitle
	
\begin{abstract}		
	Recent advances in image super-resolution (SR) explored the power of deep learning to achieve a better reconstruction performance. However, the feedback mechanism, which commonly exists in human visual system, has not been fully exploited in existing deep learning based image SR methods. In this paper, we propose an image super-resolution feedback network (SRFBN) to refine low-level representations with high-level information. Specifically, we use hidden states in an RNN with constraints to achieve such feedback manner. A feedback block is designed to handle the feedback connections and to generate powerful high-level representations. The proposed SRFBN comes with a strong early reconstruction ability and can create the final high-resolution image step by step. In addition, we introduce a curriculum learning strategy to make the network well suitable for more complicated tasks, where the low-resolution images are corrupted by multiple types of degradation. Extensive experimental results demonstrate the superiority of the proposed SRFBN in comparison with the state-of-the-art methods. Code is avaliable at \url{https://github.com/Paper99/SRFBN_CVPR19}.
\end{abstract}
	
	\section{Introduction}
	Image super-resolution (SR) is a low-level computer vision task, which aims to reconstruct a high-resolution (HR) image from its low-resolution (LR) counterpart. It is inherently ill-posed since multiple HR images may result in an identical LR image. To address this problem, numerous image SR methods have been proposed, including interpolation-based methods\cite{zhang2006edge}, reconstruction-based methods\cite{zhang2012single}, and learning-based methods~\cite{Timofte13, DBLP:journals/tip/PelegE14, Timofte15, DBLP:conf/cvpr/HuangSA15, DBLP:conf/cvpr/SchulterLB15, dong2014learning, Kim_2016_CVPR}.
	
	Since Dong \etal~\cite{dong2014learning} firstly introduced a shallow Convolutional Neural Network (CNN) to implement image SR, deep learning based methods have attracted extensive attention in recent years due to their superior reconstruction performance. The benefits of deep learning based methods mainly come from its two key factors, i.e., \textit{depth} and \textit{skip connections} (residual or dense)~\cite{Kim_2016_CVPR, Tong_2017_ICCV, Tai_2017_CVPR, Haris_2018_CVPR, Zhang_2018_CVPR, zhang2018rcan, wang2018esrgan}. The first one provides a powerful capability to represent and establish a more complex LR-HR mapping, while preserving more contextual information with larger receptive fields. The second factor can efficiently alleviate the gradient vanishing/exploding problems caused by simply stacking more layers to deepen networks.
	
	\begin{figure}[t]
		\begin{center}
			\begin{tabular}{@{}c@{}}
				\includegraphics[width=0.9\linewidth]{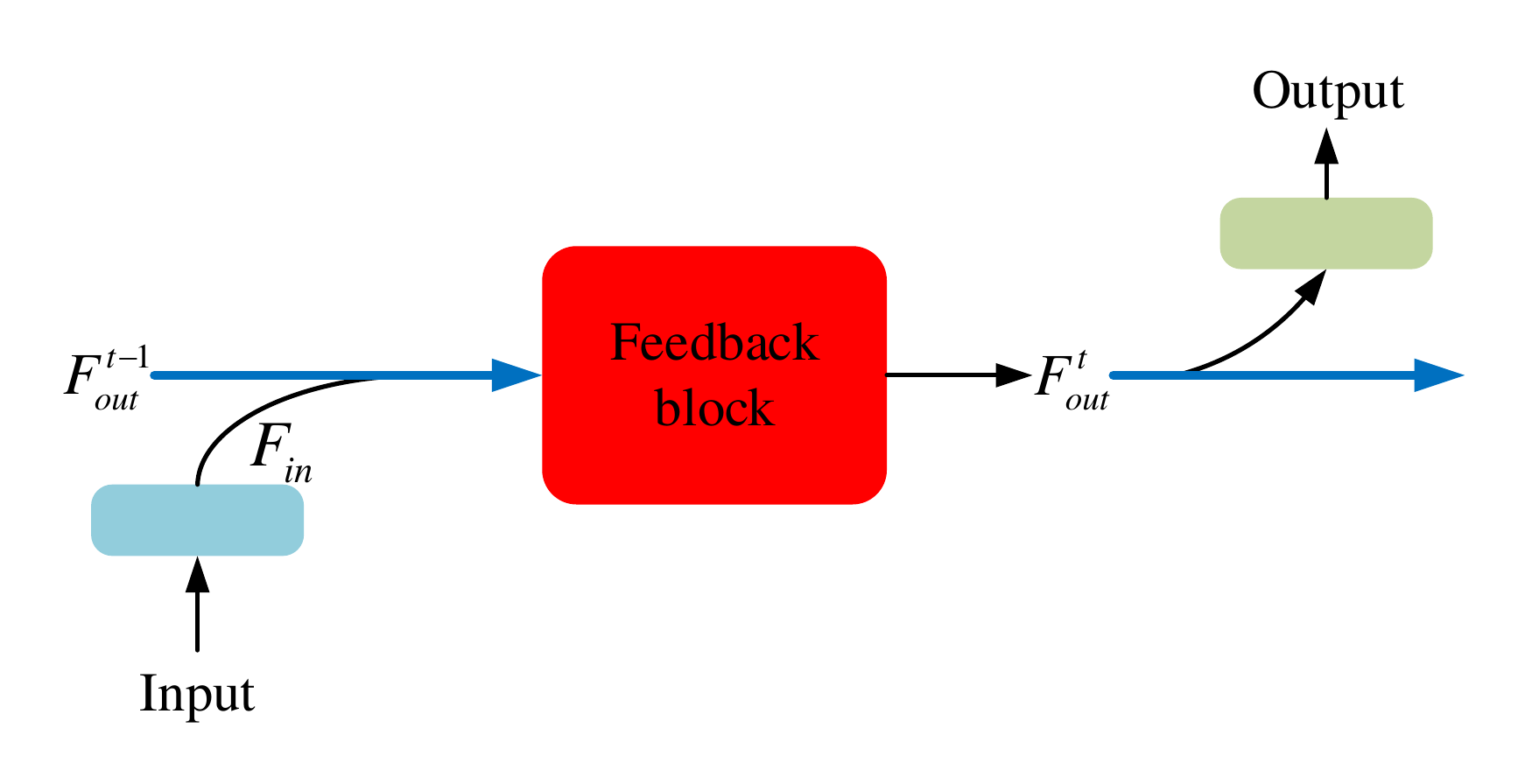}\\
				
				\small (a)
			\end{tabular}
			\begin{tabular}{@{}c@{}}
				\includegraphics[width=0.9\linewidth]{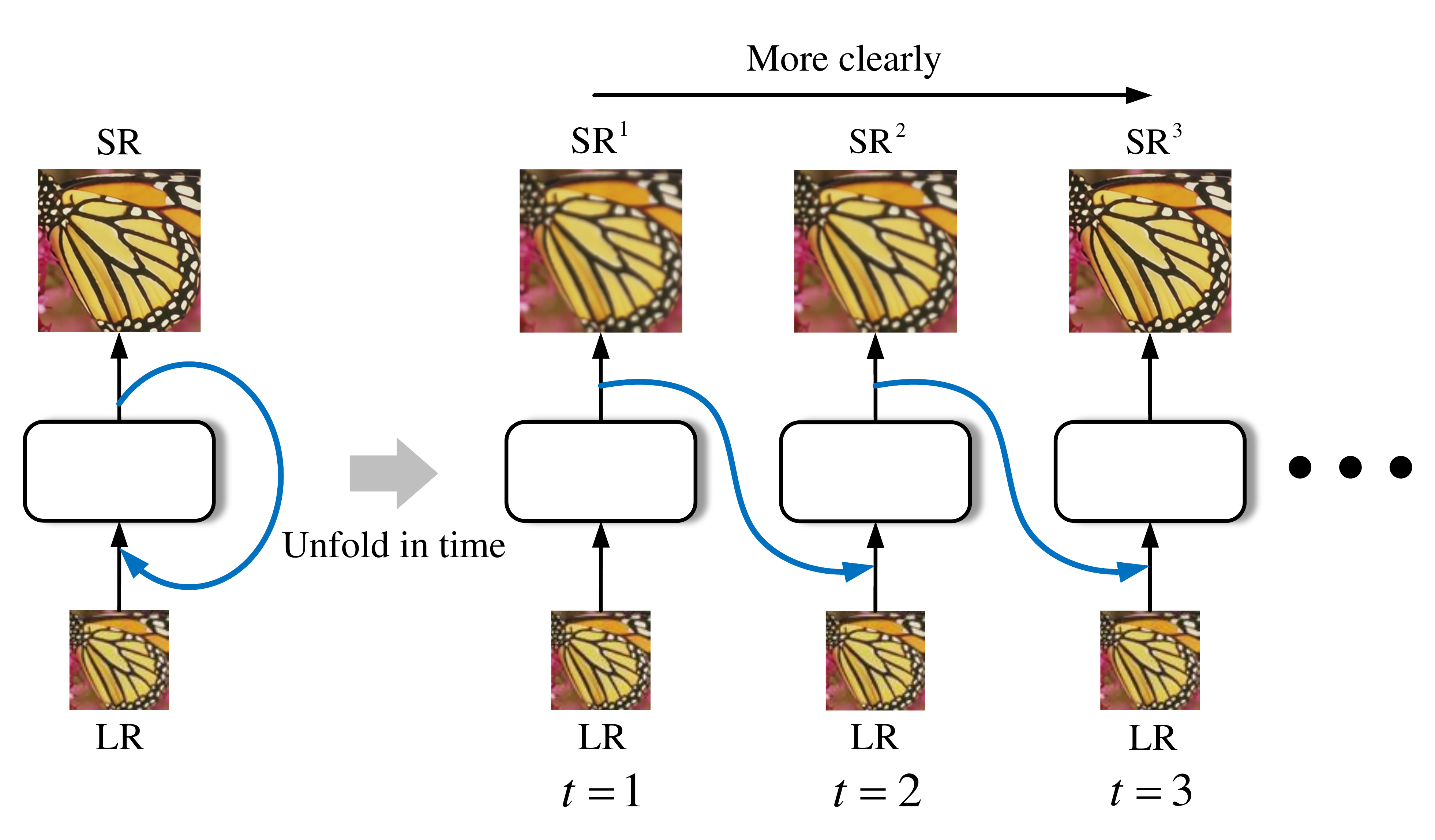}\\
				
				\small (b)
			\end{tabular}
			
		\end{center}
		\caption{The illustrations of the feedback mechanism in the proposed network. Blue arrows represent the feedback connections. (a) Feedback via the hidden state at one iteration. The feedback block (FB) receives the information of the input $F_{in}$ and hidden state from last iteration $F_{out}^{t-1}$, and then passes its hidden state $F_{out}^{t}$ to the next iteration and output. (b) The principle of our feedback scheme.}
		\label{fig:illus_fb}	
		\vspace{-0.45cm}
	\end{figure}
	
	As the depth of networks grows, the number of parameters increases. A large-capacity network will occupy huge storage resources and suffer from the overfitting problem. To reduce network parameters, the recurrent structure is often employed. Recent studies~\cite{liao2016bridging, Han_2018_CVPR} have shown that many networks with recurrent structure (\eg DRCN~\cite{Kim_2016_CVPR_DRCN} and DRRN~\cite{Tai_2017_CVPR}) can be extrapolated as a single-state Recurrent Neural Network (RNN). Similar to most conventional deep learning based methods, these networks with recurrent structure can share the information in a feedforward manner. However, the feedforward manner makes it impossible for previous layers to access useful information from the following layers, even though skip connections are employed.
	
	In cognition theory, feedback connections which link the cortical visual areas can transmit response signals from higher-order areas to lower-order areas~\cite{Hup1998Cortical, gilbert2007brain}. Motivated by this phenomenon, recent studies~\cite{NIPS2014_5276, Zamir_2017_CVPR} have applied the feedback mechanism to network architectures. The feedback mechanism in these architectures works in a top-down manner, carrying high-level information back to previous layers and refining low-level encoded information.
	
	In this paper, we propose a novel network for image SR, namely the Super-Resolution Feedback Network (SRFBN), in order to refine low-level information using high-level one through feedback connections. The proposed SRFBN is essentially an RNN with a feedback block (FB), which is specifically designed for image SR tasks. The FB is constructed by multiple sets of up- and down-sampling layers with dense skip connections to generate powerful high-level representations. Inspired by~\cite{Zamir_2017_CVPR}, we use the output of the FB, i.e., a hidden state in an unfolded RNN, to achieve the feedback manner (see Fig.~\ref{fig:illus_fb}(a)). The hidden state at each iteration flows into the next iteration to modulate the input. To ensure the hidden state contains the information of the HR image, we connect the loss to each iteration during the training process. The principle of our feedback scheme is that the information of a coarse SR image can facilitate an LR image to reconstruct a better SR image (see Fig.~\ref{fig:illus_fb}(b)). Furthermore, we design a curriculum for the case, in which the LR image is generated by a complex degradation model. For each LR image, its target HR images for consecutive iterations are arranged from easy to hard based on the recovery difficulty. Such curriculum learning strategy well assists our proposed SRFBN in handling complex degradation models. Experimental results demonstrate the superiority of our proposed SRFBN against other state-of-the-art methods.
		
	In summary, our main contributions are as follows:
	
	\begin{itemize}
		\setlength{\itemsep}{0pt}
		\setlength{\parsep}{0pt}
		\setlength{\parskip}{0pt}
		\item Proposing an image super-resolution feedback network (SRFBN), which employs a feedback mechanism. High-level information is provided in top-down feedback flows through feedback connections. Meanwhile, such recurrent structure with feedback connections provides strong early reconstruction ability, and requires only few parameters.
		\item Proposing a feedback block (FB), which not only efficiently handles feedback information flows, but also enriches high-level representations via up- and down-sampling layers, and dense skip connections. 
		\item Proposing a curriculum-based training strategy for the proposed SRFBN, in which HR images with increasing reconstruction difficulty are fed into the network as targets for consecutive iterations. This strategy enables the network to learn complex degradation models step by step, while the same strategy is impossible to settle for those methods with only one-step prediction.
	\end{itemize}
	
	\section{Related Work}
	\subsection{Deep learning based image super-resolution}
	
	Deep learning has shown its superior performance in various computer vision tasks including image SR. Dong \etal~\cite{dong2016image} firstly introduced a three-layer CNN in image SR to learn a complex LR-HR mapping. Kim \etal~\cite{Kim_2016_CVPR} increased the depth of CNN to 20 layers for more contextual information usage in LR images. In \cite{Kim_2016_CVPR}, a skip connection was employed to overcome the difficulty of optimization when the network became deeper. Recent studies have adopted different kind of skip connections to achieve remarkable improvement in image SR. SRResNet\cite{DBLP:conf/cvpr/LedigTHCCAATTWS17} and EDSR\cite{lim2017enhanced} applied residual skip connections from \cite{He2016Deep}. SRDenseNet\cite{Tong_2017_ICCV} applied dense skip connections from \cite{Huang2016Densely}. Zhang \etal~\cite{Zhang_2018_CVPR} combined local/global residual and dense skip connections in their RDN. Since the skip connections in these network architectures use or combine hierarchical features \textit{in a bottom-up way}, the low-level features can only receive the information from previous layers, lacking enough contextual information due to the limitation of small receptive fields. These low-level features are reused in the following layers, and thus further restrict the reconstruction ability of the network. To fix this issue, we propose a super-resolution feedback network (SRFBN), in which high-level information flows through feedback connections \textit{in a top-down manner} to correct low-level features using more contextual information.
	
	Meanwhile, with the help of skip connections, neural networks go deeper and hold more parameters. Such large-capacity networks occupy huge amount of storage resources and suffer from overfitting. To effectively reduce network parameters and gain better generalization power, the recurrent structure was employed\cite{Kim_2016_CVPR_DRCN, Tai_2017_CVPR, tai2017memnet}. Particularly, the recurrent structure plays an important role to realize the feedback process in the proposed SRFBN (see Fig.~\ref{fig:illus_fb}(b)). 
	
	\begin{figure*}
		\centering
		\includegraphics[width=0.9\textwidth]{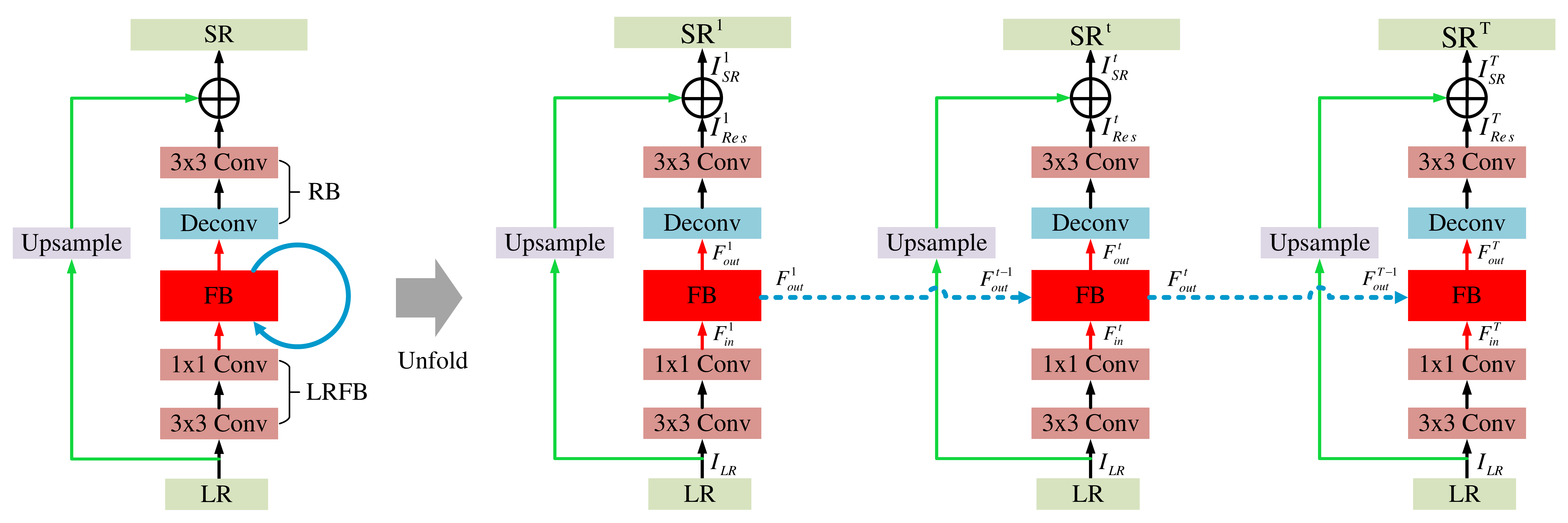}
		\caption{The architecture of our proposed super-resolution feedback network (SRFBN). Blue arrows represent feedback connections. Green arrows represent global residual skip connections.}
		\label{SRFBN}
	\end{figure*}
	
	\subsection{Feedback mechanism}
	The feedback mechanism allows the network to carry a notion of output to correct previous states. Recently, the feedback mechanism has been adopted by many network architectures for various vision tasks\cite{Carreira2015Human, DBLP:conf/iccv/CaoLYYWWHWHXRH15, Zamir_2017_CVPR, Haris_2018_CVPR, Han_2018_CVPR, DBLP:conf/aaai/SamB18}.
	
	For image SR, a few studies also showed efforts to introduce the feedback mechanism. Based on back-projection, Haris \etal~\cite{Haris_2018_CVPR} designed up- and down-projection units to achieve iterative error feedback. Han \etal~\cite{Han_2018_CVPR} applied a delayed feedback mechanism which transmits the information between two recurrent states in a dual-state RNN. However, the flow of information from the LR image to the final SR image is still feedforward in their network architectures unlike ours. 
	
	The most relevant work to ours is \cite{Zamir_2017_CVPR}, which transfers the hidden state with high-level information to the information of an input image to realize feedback in an convolutional recurrent neural network. However, it aims at solving high-level vision tasks, \eg classification. To fit a feedback mechanism in image SR, we elaborately design a feedback block (FB) as the basic module in our SRFBN, instead of using ConvLSTM as in \cite{Zamir_2017_CVPR}. The information in our FB efficiently flows across hierarchical layers through dense skip connections. Experimental results indicate our FB has superior reconstruction performance than ConvLSTM\footnote{Further analysis can be found in our supplementary material.} and thus is more suitable for image SR tasks.
	
	\subsection{Curriculum learning}
	Curriculum learning\cite{Bengio2009Curriculum}, which gradually increases the difficulty of the learned target, is well known as an efficient strategy to improve the training procedure. Early work of curriculum learning mainly focuses on a single task. Pentina \etal~\cite{Pentina2014Curriculum} extended curriculum learning to multiple tasks in a sequential manner. Gao \etal~\cite{Gao2017On} utilized curriculum learning to solve the fixation problem in image restoration. Since their network is limited to a one-time prediction, they enforce a curriculum through feeding different training data in terms of the complexity of tasks as epoch increases during the training process. In the context of image SR, Wang \etal~\cite{Wang2018A} designed a curriculum for the pyramid structure, which gradually blends a new level of the pyramid in previously trained networks to upscale an LR image to a bigger size.
	
	While previous works focus on a single degradation process, we enforce a curriculum to the case, where the LR image is corrupted by multiple types of degradation. The curriculum containing easy-to-hard decisions can be settled for one query to gradually restore the corrupted LR image.
	
	\section{Feedback Network for Image SR}
	\label{3.1}
	Two requirements are contained in a feedback system: (1) iterativeness and (2) rerouting the output of the system to correct the	input in each loop. Such iterative cause-and-effect process helps to achieve the principle of our feedback scheme for	image SR: high-level information can guide an LR image to recover a better SR image (see Fig.~\ref{fig:illus_fb}(b)). In the proposed	network, there are three indispensable parts to enforce our feedback scheme: (1) tying the loss at each iteration (to force the network to reconstruct an SR image at each iteration and thus allow the hidden state to carry a notion of high-level information), (2) using recurrent structure (to achieve iterative process) and (3) providing an LR input at each iteration (to ensure the availability of low-level information, which is needed to be refined). Any absence of these three parts will fail the network to drive the feedback flow.
	\subsection{Network structure}
	As shown in Fig.~\ref{SRFBN}, our proposed SRFBN can be unfolded to $T$ iterations, in which each iteration $t$ is temporally ordered from $1$ to $T$. In order to make the hidden state in SRFBN carry a notion of output, we tie the loss for every iteration. The description of the loss function can be found in Sec.~\ref{3.3}. The sub-network placed in each iteration $t$ contains three parts: an LR feature extraction block (LRFB), a feedback block (FB) and a reconstruction block (RB). The weights of each block are shared across time. The global residual skip connection at each iteration $t$ delivers an upsampled image to bypass the sub-network. Therefore, the purpose of the sub-network at each iteration $t$ is to recover a residual image $I_{Res}^{t}$ while input a low-resolution image $I_{LR}$. We denote $Conv(s,n)$ and $Deconv(s,n)$ as a convolutional layer and a deconvolutional layer respectively, where $s$ is the size of the filter and $n$ is the number of filters. 
	
	The LR feature extraction block consists of $Conv(3,4m)$ and $Conv(1,m)$. $m$ denotes the base number of filters. We provide an LR input $I_{LR}$ for the LR feature extraction block, from which we obtain the shallow features $F_{in}^{t}$ containing the information of an LR image:
	\begin{equation}
	F_{in}^{t} = f_{LRFB}(I_{LR}),
	\end{equation}
	where $f_{LRFB}$ denotes the operations of the LR feature extraction block. $F_{in}^{t}$ are then used as the input to the FB. In addition, $F_{in}^{1}$ are regarded as the initial hidden state $F_{out}^{0}$.
	
	The FB at the $t$-th iteration receives the hidden state from previous iteration $F_{out}^{t-1}$ through a feedback connection and shallow features $F_{in}^{t}$. $F_{out}^{t}$ represents the output of the FB. The mathematical formulation of the FB is:	
	\begin{equation}
	F_{out}^{t} = f_{FB}(F_{out}^{t-1},F_{in}^{t}),
	\end{equation}
	where $f_{FB}$ denotes the operations of the FB and actually represents the feedback process as shown in Fig.~\ref{fig:illus_fb}(b). More details of the FB can be found in Sec.~\ref{Sec3.2}. 
	
	 The reconstruction block uses $Deconv(k,m)$ to upscale LR features $F_{out}^{t}$ to HR ones and $Conv(3,c_{out})$ to generate a residual image $I_{Res}^{t}$. The mathematical formulation of the reconstruction block is:
	\begin{equation}
	I_{Res}^{t} = f_{RB}(F_{out}^{t}),
	\end{equation}
	where $f_{RB}$ denotes the operations of the reconstruction block. 
	
	The output image $I_{SR}^{t}$ at the $t$-th iteration can be obtained by:
	\begin{equation}
	I_{SR}^{t} = I_{Res}^{t}+f_{UP}(I_{LR}),
	\end{equation}
	where $f_{UP}$ denotes the operation of an upsample kernel. The choice of the upsample kernel is arbitrary. We use a bilinear upsample kernel here.
	After $T$ iterations, we will get totally $T$ SR images $(I_{SR}^{1}, I_{SR}^{2}, ..., I_{SR}^{T})$.  
	
	\subsection{Feedback block}
	\label{Sec3.2}
	
	\begin{figure}[h]
		\centering
		\includegraphics[width=.45\textwidth]{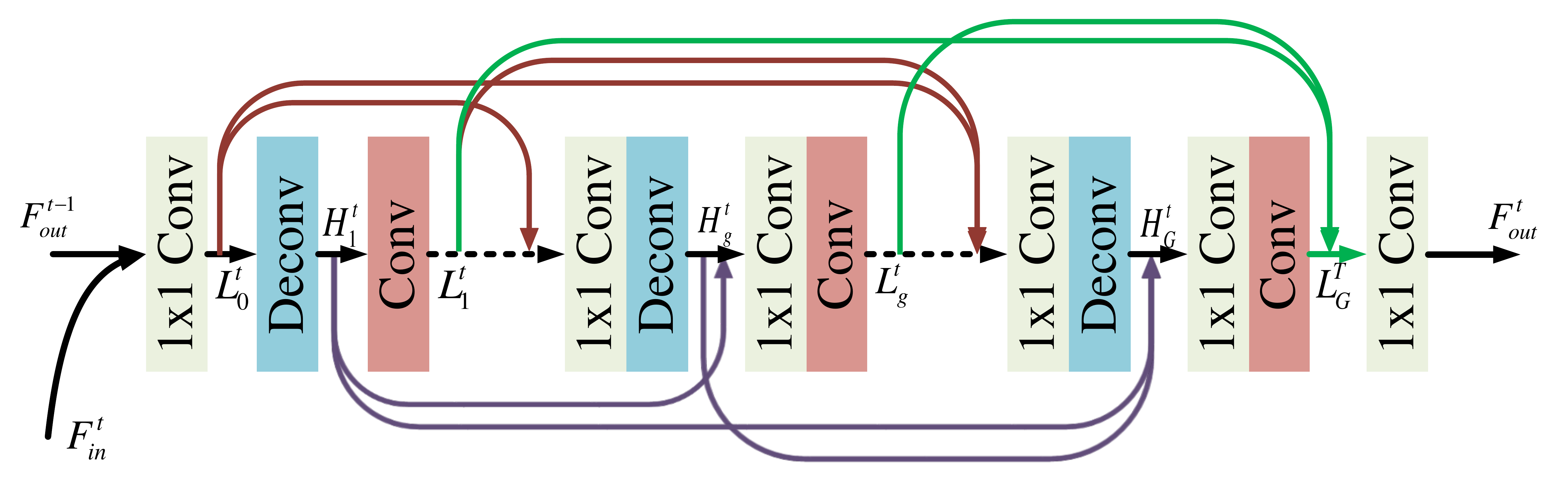}
		\caption{Feedback block (FB).}
		\label{FB}
	\end{figure}
	
    As shown in Fig.~\ref{FB}, the FB at the $t$-th iteration receives the feedback information $F_{out}^{t-1}$ to correct low-level representations $F_{in}^{t}$, and then passes more powerful high-level representations $F_{out}^{t}$ to the next iteration and the reconstruction block. The FB contains G projection groups sequentially with dense skip connections among them. Each projection group, which can project HR features to LR ones, mainly includes an upsample operation and a downsample operation.
	
	At the beginning of the FB, $F_{in}^{t}$ and $F_{out}^{t-1}$ are concatenated and compressed by $Conv(1,m)$ to refine input features $F_{in}^{t}$ by feedback information $F_{out}^{t-1}$, producing the refined input features $L_{0}^{t}$:
	\begin{equation}
	L_{0}^{t} = C_{0}([F_{out}^{t-1}, F_{in}^{t}]),
	\end{equation}
	where $C_{0}$ refers to the initial compression operation and $[F_{out}^{t-1}, F_{in}^{t}]$ refers to the concatenation of $F_{out}^{t-1}$ and $F_{in}^{t}$. Let $H_{g}^{t}$ and $L_{g}^{t}$ be the HR and LR feature maps given by the $g$-th projection group in the FB at the $t$-th iteration. $H_{g}^{t}$ can be obtained by:
	\begin{equation}
	H_{g}^{t} = C_{g}^{\uparrow}([L_{0}^{t}, L_{1}^{t}, ..., L_{g-1}^{t}]),
	\end{equation}
	where $C_{g}^{\uparrow}$ refers to the upsample operation using $Deconv(k,m)$ at the $g$-th projection group. Correspondingly, $L_{g}^{t}$ can be obtained by
	\begin{equation}
	L_{g}^{t} = C_{g}^{\downarrow}([H_{1}^{t}, H_{2}^{t}, ..., H_{g}^{t}]),
	\end{equation}
	where $C_{g}^{\downarrow}$ refers to the downsample operation using $Conv(k,m)$ at the $g$-th projection group.
	Except for the first projection group, we add $Conv(1,m)$ before $C_{g}^{\uparrow}$ and $C_{g}^{\downarrow}$ for parameter and computation efficiency.
	
	In order to exploit useful information from each projection group and map the size of input LR features $F_{in}^{t+1}$ at the next iteration, we conduct the feature fusion (green arrows in Fig.~\ref{FB}) for LR features generated by projection groups to generate the output of FB:
	\begin{equation}
	F_{out}^{t} = C_{FF}([L_{1}^{t}, L_{2}^{t},...,L_{G}^{t}]),
	\end{equation}
	where $C_{FF}$ represents the function of $Conv(1,m)$.
	
	\subsection{Curriculum learning strategy}
	\label{3.3}
	We choose $L1$ loss to optimize our proposed network. T target HR images $(I_{HR}^{1}, I_{HR}^{2}, ..., I_{HR}^{T})$ are placed to fit in the multiple output in our proposed network. $(I_{HR}^{1}, I_{HR}^{2}, ..., I_{HR}^{T})$ are identical for the single degradation model. For complex degradation models, $(I_{HR}^{1}, I_{HR}^{2}, ..., I_{HR}^{T})$ are ordered based on the difficulty of tasks for T iterations to enforce a curriculum. The loss function in the network can be formulated as:
	\begin{equation}
	L(\Theta) = \frac{1}{T}\sum_{t=1}^{T}{W^t}\left \|I_{HR}^{t} - I_{SR}^{t}\right \|_{1},
	\end{equation}
	where $\Theta$ denotes to the parameters of our network. $W^{t}$ is a constant factor which demonstrates the worth of the output at the $t$-th iterations. As \cite{Zamir_2017_CVPR} do, we set the value to 1 for each iteration, which represents each output has equal contribution. Details about settings of target HR images for complex degradation models will be revealed in Sec.~\ref{4.4}.
	
	\subsection{Implementation details}
	We use PReLU\cite{DBLP:conf/iccv/HeZRS15} as the activation function following all convolutional and deconvolutional layers except the last layer in each sub-network. Same as \cite{Haris_2018_CVPR}, we set various $k$ in $Conv(k,m)$ and $Deconv(k,m)$ for different scale factors to perform up- and down-sampling operations. For $\times2$ scale factor, we set $k$ in $Conv(k,m)$ and $Deconv(k,m)$ as 6 with two striding and two padding. Then, for $\times3$ scale factor, we set $k=7$  with three striding and two padding. Finally, for $\times4$ scale factor, we set $k=8$ with four striding and two padding. We take the SR image $I_{SR}^T$ at the last iteration as our final SR result unless we specifically analysis every output image at each iteration. Our network can process both gray and color images, so $c_{out}$ can be 1 or 3 naturally.
	
	\section{Experimental Results}
	\label{exp}
	
	\subsection{Settings}
	\label{4.1}
	\textbf{Datasets and metrics.} We use DIV2K\cite{Agustsson_2017_CVPR_Workshops} and Flickr2K as our training data. To make full use of data, we adopt data augmentation as \cite{lim2017enhanced} do. We evaluate SR results under PSNR and SSIM\cite{DBLP:journals/tip/WangBSS04} metrics on five standard benchmark datasets: Set5\cite{DBLP:conf/bmvc/BevilacquaRGA12}, Set14\cite{ZeydeEP10}, B100\cite{DBLP:conf/iccv/MartinFTM01}, Urban100\cite{DBLP:conf/cvpr/HuangSA15}, and Manga109\cite{DBLP:journals/mta/MatsuiIAFOYA17}. To keep consistency with previous works, quantitative results are only evaluated on luminance (Y) channel.
		
	\textbf{Degradation models.} In order to make fair comparison with existing models, we regard bicubic downsampling as our standard degradation model (denoted as \textbf{BI}) for generating LR images from ground truth HR images. To verify the effectiveness of our curriculum learning strategy, we further conduct two experiments involving two other multi-degradation models as \cite{Zhang_2018_CVPR} do in Sec.~\ref{4.4} and \ref{4.5.3}. We define \textbf{BD} as a degradation model which applies Gaussian blur followed by downsampling to HR images. In our experiments, we use 7x7 sized Gaussian kernel with standard deviation 1.6 for blurring. Apart from the \textbf{BD} degradation model, \textbf{DN} degradation model is defined as bicubic downsampling followed by adding Gaussian noise, with noise level of 30. 
	
	\begin{table}[!htbp]
		\centering
		\resizebox{.40\textwidth}{!}{\begin{tabular}{|c|c|c|c|}
				\hline
				Scale factor     & $\times2$    & $\times3$    & $\times4$    \\
				\hline\hline
				Input patch size & $60\times60$ & $50\times50$ & $40\times40$ \\
				\hline
		\end{tabular}}	
		\smallskip
		\caption{The settings of input patch size.\label{input_size}}
	\end{table}
	
	\textbf{Training settings.} We train all networks with the batch-size of 16. To fully exploit contextual information from LR images, we feed RGB image patches with different patch size based on the upscaling factor. The settings of input patch size are listed in Tab.~\ref{input_size}. The network parameters are initialized using the method in \cite{DBLP:conf/iccv/HeZRS15}. Adam\cite{Kingma2014Adam} is employed to optimize the parameters of the network with initial learning rate 0.0001. The learning rate multiplies by 0.5 for every 200 epochs. We implement our networks with Pytorch framework and train them on NVIDIA 1080Ti GPUs.
	
	\begin{figure}[!htbp]
		\begin{center}
			\begin{tabular}{@{}c@{}c@{}}
				\includegraphics[width=0.5\linewidth]{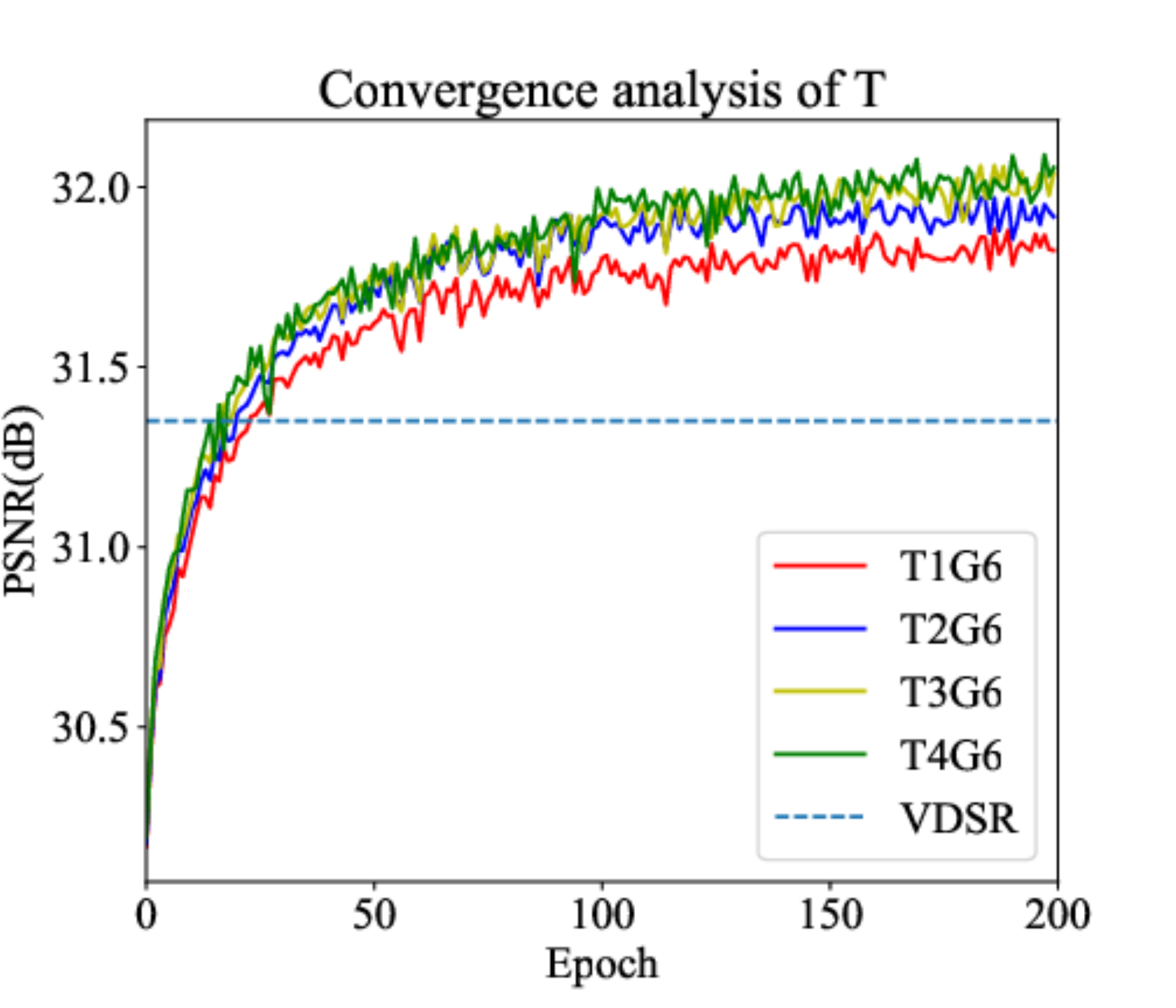} & \includegraphics[width=0.5\linewidth]{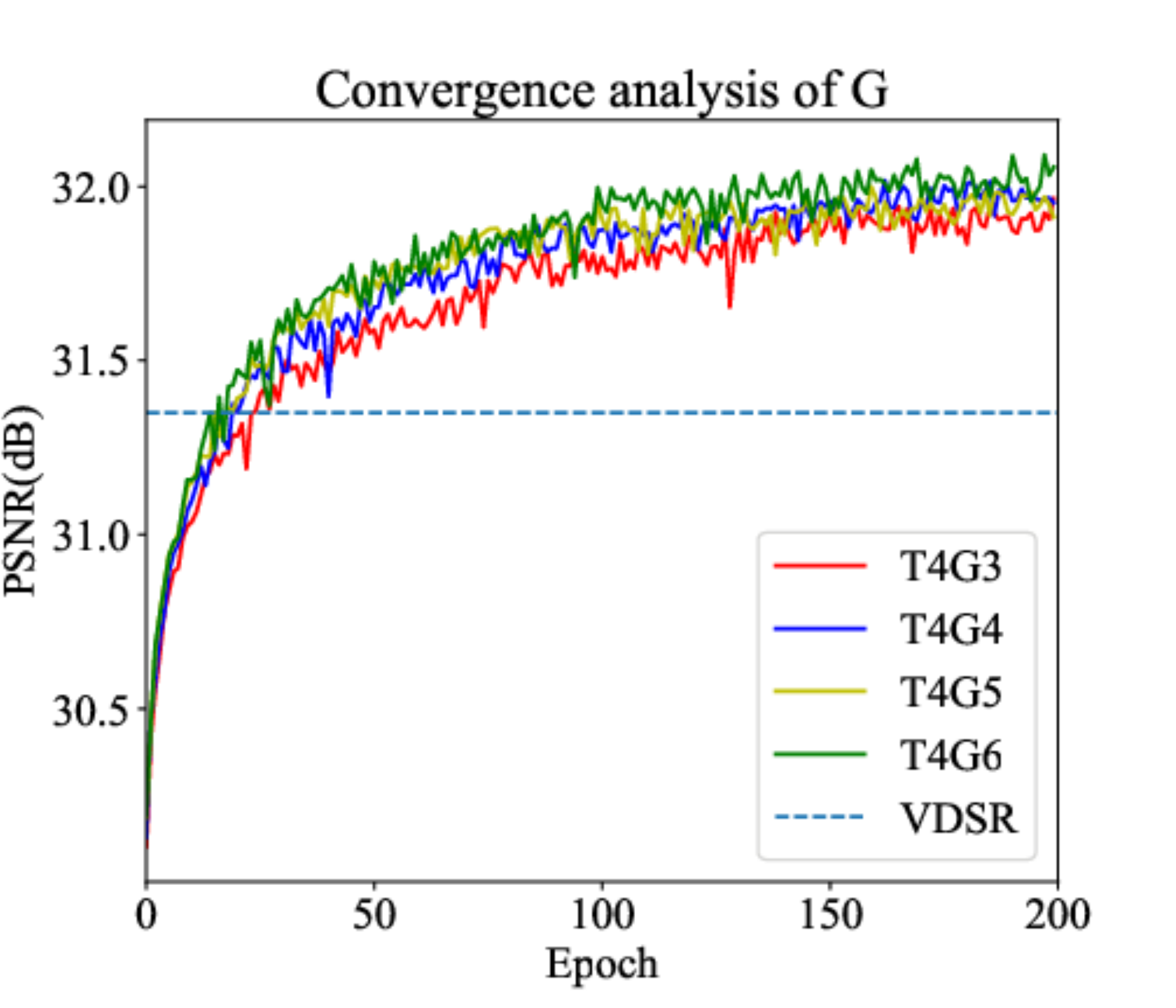}\\
				
				\small (a)& \small (b)
			\end{tabular}
			
		\end{center}
		\caption{Convergence analysis of T and G on Set5 with scaling factor $\times 4$.}
		\label{T_and_G}	
		\vspace{-0.45cm}
	\end{figure}

	\subsection{Study of T and G}   
	 In this subsection, we explore the influence of the number of iterations (denoted as T) and the number of projection groups in the feedback block (denoted as G). The base number of filters $m$ is set to 32 in subsequent experiments. We first investigate the influence of T by fixing G to 6. It can be observed from Fig.~\ref{T_and_G}(a) that with the help of feedback connection(s), the reconstruction performance is significantly improved compared with the network without feedback connections (T=1). Besides, as T continues to increase, the reconstruction quality keeps rising. In other words, our feedback block surely benefits the information flow across time. We then study the influence of G by fixing T to 4. Fig.~\ref{T_and_G}(b) shows that larger G leads to higher accuracy due to stronger representative ability of deeper networks. In conclusion, choosing larger T or G contributes to better results. It is worth noticing that small T and G still outperform VDSR\cite{Kim_2016_CVPR}. In the following discussions, we use SRFBN-L (T=4, G=6) for analysis. 
	
	\begin{table}[!htbp]
		\centering
		\begin{tabular}{|c|c|c|c|c|}
			\hline
			No. Prediction & 1st & 2nd    & 3rd    & 4th    \\ \hline\hline
			SRFBN-L-FF &  30.69  & 31.74 & 32.00 & 32.09 \\ \hline
			SRFBN-L    &  \textbf{31.85}  & \textbf{32.06} & \textbf{32.11} & \textbf{32.11} \\ \hline
		\end{tabular}%
		\smallskip
		\caption{The impact of feedback on Set5 with scale factor $\times4$.\label{ff_vs_fb}}
		\vspace{-0.45cm}			
	\end{table}
	
	\subsection{Feedback vs. feedforward}
	To investigate the nature of the feedback mechanism in our network, we compare the feedback network with feedforward one in this subsection.
	
	We first demonstrate the superiority of the feedback mechanism over its feedforward counterpart. By simply disconnecting the loss to all iterations except the last one, the network is thus impossible to reroute a notion of output to low-level representations and is then degenerated to a feedforward one (however still retains its recurrent property), denoted as SRFBN-L-FF. SRFBN-L and SRFBN-L-FF both have four iterations, producing four intermediate output. We then compare the PSNR values of all intermediate SR images from both networks. The results are shown in Tab.~\ref{ff_vs_fb}. SRFBN-L outperforms SRFBN-L-FF at every iteration, from which we conclude that the feedback network is capable of producing high quality early predictions in contrast to feedforward network. The experiment also indicates that our proposed SRFBN does benefit from the feedback mechanism, instead of only rely on the power of the recurrent structure. Except for the above discussions about the necessity of early losses, we also conduct two more abalative experiments to verify other parts (discussed in Sec.~\ref{3.1}) which form our feedback system. By turning off weights sharing across iterations, the PSNR value in the proposed network is decreased from 32.11dB to 31.82dB on Set5 with scale factor $\times 4$. By disconnecting the LR input at each iteration except the first iteration, the PSNR value is decreased by 0.17dB. 
	\vspace{-0.4cm}
	\begin{figure}[!htbp]
		\centering
		\includegraphics[width=.5\textwidth]{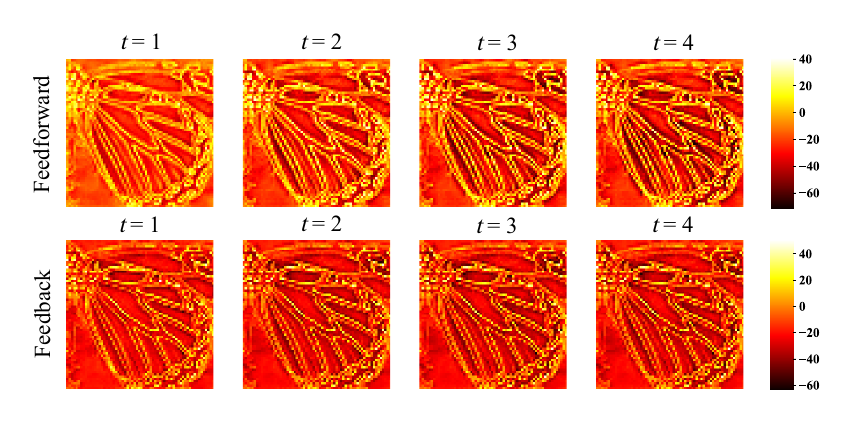}
		\caption{Average feature maps of feedforward and feedback networks.}
		\label{featuremap}
	\end{figure}

	To dig deeper into the difference between feedback and feedforward networks, we visualize the average feature map of every iteration in SRFBN-L and SRFBN-L-FF, illustrated in Fig.~\ref{featuremap}. Each average feature map is the mean of $F_{out}^{t}$ in channel dimension, which roughly represents the output of the feedback block at the $t$-th iteration. Our network with global residual skip connections aims at recovering the residual image. In other words, the tasks of our network are to suppress the smooth area of the original input image\cite{Hui-IDN-2018} and to predict high-frequency components (\ie edges and contours). From Fig.~\ref{featuremap}, we have two observations. First, compared with the feedforward network at early iterations, feature maps acquired from the feedback network contain more negative values, showing a stronger effect of suppressing the smooth area of the input image, which further leads to a more accurate residual image. To some extent, this illustration reflects the reason why the feedback network has more powerful early reconstruction ability than the feedforward one. The second observation is that the feedback network learns different representations in contrast to feedforward one when handling the same task. In the feedforward network, feature maps vary significantly from the first iteration ($t$=1) to the last iteration ($t$=4): the edges and contours are outlined at early iterations and then the smooth areas of the original image are suppressed at latter iterations. The distinct patterns demonstrate that the feedforward network forms a hierarchy of information through layers, while the feedback network is allowed to devote most of its efforts to take a self-correcting process, since it can obtain well-developed feature representations at the initial iteration. This further indicates that $F_{out}^{t}$ containing high-level information at the $t$-th iteration in the feedback network will urge previous layers at subsequent iterations to generate better representations.
	\begin{table}[!htbp]
		\begin{center}
			\resizebox{.40\textwidth}{!}{\begin{tabular}{|c|c|c|c|c|}
					\hline
					\multirow{2}{*}{Model} & \multicolumn{2}{c|}{from scratch} & \multicolumn{2}{c|}{from pretrained} \\ \cline{2-5} 
					& w/o CL          & with CL         & w/o CL    & with CL           \\ \hline \hline
					\textbf{BD}                           & 29.78           & 29.96           & 29.98     & \textbf{30.03}    \\
					\textbf{DN}                           &     26.92            & 26.93                &  26.96	         & \textbf{26.98}                  \\ \hline
					
			\end{tabular}}
			\smallskip
			\caption{The investigation of curriculum learning (CL) on \textbf{BD} and \textbf{DN} degradation models with scale factor $\times4$. The average PSNR values are evaluated on Set5.}
			\label{CL_table}
		\end{center}
		\vspace{-1cm}	
	\end{table}	

	\begin{figure}[!htbp]
		\centering
		\includegraphics[width=.45\textwidth]{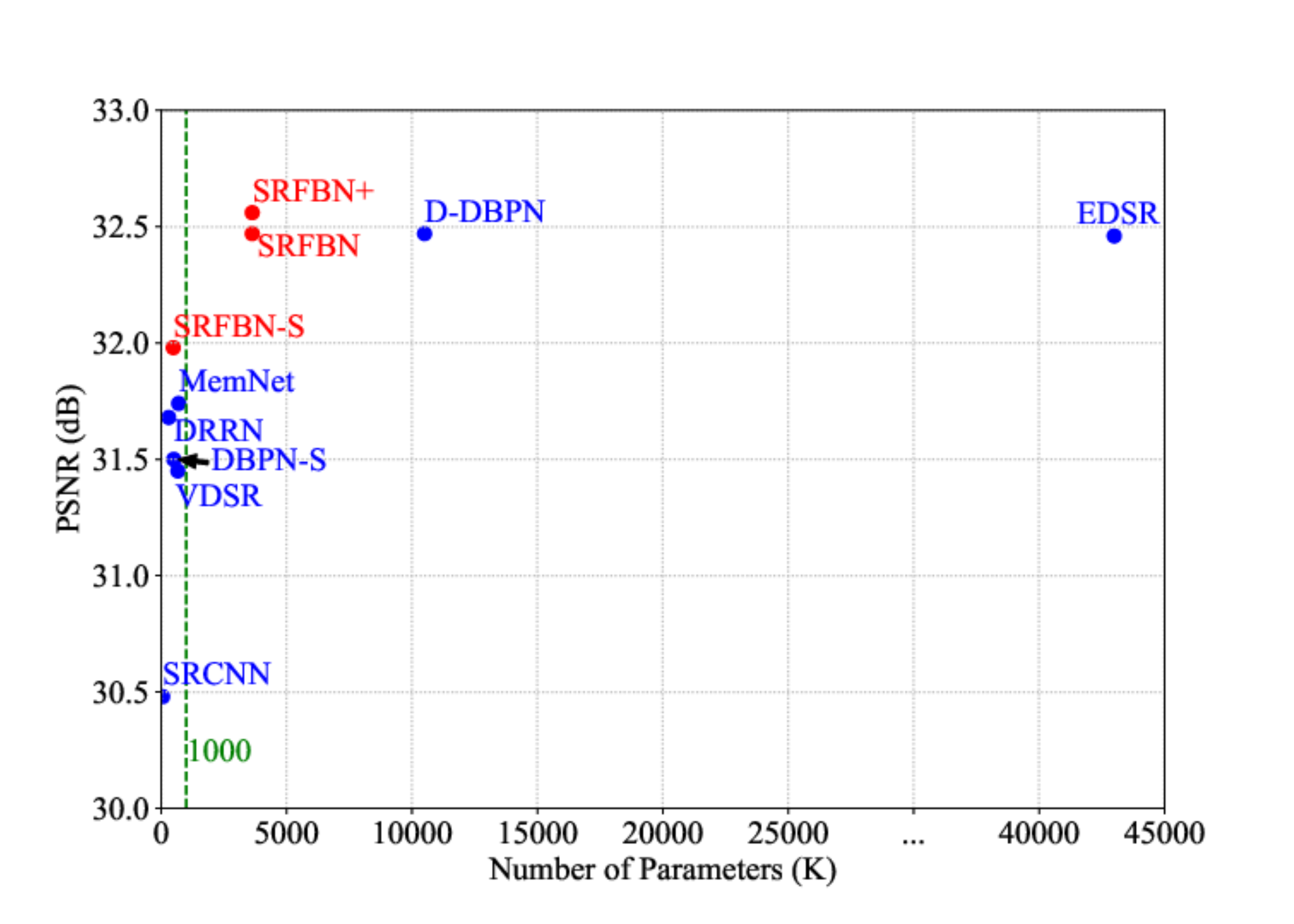}
		\caption{Performance and number of parameters. Results are evaluated on Set5 with scale factor $\times4$. Red points represent our proposed networks.}
		\label{num_param}
		\vspace{-0.45cm}
	\end{figure}
	\subsection{Study of curriculum learning}
	\label{4.4}
	As mentioned in Sec.~\ref{4.1}, we now present our results for two experiments on two different degradation models, \ie \textbf{BD} and \textbf{DN}, to show the effectiveness of our curriculum learning strategy.
	
	We formulate the curriculum based on the recovery difficulty. For example, to guide the network to learn recovering a \textbf{BD} operator corrupted image step by step, we provide a Gaussian blurred HR image as (intermediate) ground truth so that the network only needs to learn the inversion of a single downsampling operator at early iterations. Original HR image is provided at latter iterations as a senior challenge. Specifically, we empirically provide blurred HR images at first two iterations and original HR images at remaining two iterations for experiments with the \textbf{BD} degradation model. For experiments with the \textbf{DN} degradation model, we instead use noisy HR images at first two iterations and HR images without noise at last two iterations. 

	We also examine the compatibility of this strategy with two common training processes, \ie training from scratch and fine-tuning on a network pretrained on the \textbf{BI} degradation model. The results shown in Tab.~\ref{CL_table} infer that the curriculum learning strategy well assists our proposed SRFBN in handling \textbf{BD} and \textbf{DN} degradation models under both circumstances. We also observe that fine-tuning on a network pretrained on the \textbf{BI} degradation model leads to higher PSNR values than training from scratch. 

	\begin{table*}[!htbp]	
	\centering	
	\resizebox{\textwidth}{!}{\begin{tabular}{|c|c|c|c|c|c|c|c||c|c|c|c|}
			\hline

			\multirow{2}{*}{Dataset}  & \multirow{2}{*}{Scale} & \multirow{2}{*}{Bicubic} & SRCNN        & VDSR & DRRN        & MemNet& SRFBN-S & EDSR & D-DBPN & SRFBN & SRFBN+ \\
			&                        &                          &   \cite{dong2016image}       &  \cite{Kim_2016_CVPR}    &    \cite{Tai_2017_CVPR}       &   \cite{tai2017memnet} & (Ours)     &  \cite{lim2017enhanced}   &  \cite{Haris_2018_CVPR}    &   (Ours)  & (Ours)     \\ \hline\hline
			\multirow{3}{*}{Set5}     & $\times2$ & 33.66/0.9299 & 36.66/0.9542 &  37.53/0.9590  & 37.74/0.9591    &  37.78/0.9597 & 37.78/0.9597     &  {\color{blue} 38.11}/0.9602  &   38.09/0.9600   &   {\color{blue} 38.11}/{\color{blue} 0.9609}  &  {\color{red} 38.18}/{\color{red} 0.9611}	     \\  
			& $\times3$ & 30.39/0.8682 & 32.75/0.9090 &  33.67/0.9210    & 34.03/0.9244        &   34.09/0.9248 & 34.20/0.9255   &  34.65/0.9280    &  -/-   &   {\color{blue} 34.70}/{\color{blue} 0.9292}   &   {\color{red} 34.77}/{\color{red} 0.9297}    \\  
			& $\times4$ & 28.42/0.8104 & 30.48/0.8628 &  31.35/0.8830    & 31.68/0.8888  &   31.74/0.8893  & 31.98/0.8923    &  32.46/0.8968    &  {\color{blue} 32.47}/0.8980    &  {\color{blue} 32.47}/{\color{blue} 0.8983}    &   {\color{red} 32.56}/{\color{red} 0.8992}     \\ \hline\hline
			\multirow{3}{*}{Set14}    & $\times2$ &    30.24/0.8688          &   32.45/0.9067            &   33.05/0.9130   &     33.23/0.9136    &  33.28/0.9142   & 33.35/0.9156    &   {\color{red} 33.92}/0.9195    &  33.85/0.9190    &  33.82/{\color{blue} 0.9196}    & {\color{blue} 33.90}/{\color{red} 0.9203}     \\  
			& $\times3$ &     27.55/0.7742         &     29.30/0.8215         &   29.78/0.8320   &      29.96/0.8349        &    30.00/0.8350 & 30.10/0.8372   &   {\color{blue} 30.52}/{\color{blue} 0.8462}   &   -/-   &  30.51/0.8461    &  {\color{red} 30.61}/{\color{red} 0.8473}     \\ 
			& $\times4$ &     26.00/0.7027         & 27.50/0.7513   &  28.02/0.7680    &   28.21/0.7721    &    28.26/0.7723  & 28.45/0.7779      &   28.80/{\color{blue} 0.7876}     &   {\color{blue} 28.82}/0.7860   &  28.81/0.7868      &    {\color{red} 28.87}/{\color{red} 0.7881}      \\ \hline\hline
			\multirow{3}{*}{B100}     & $\times2$ &  29.56/0.8431            &     31.36/0.8879        &   31.90/0.8960   &     32.05/0.8973     &    32.08/0.8978  & 32.00/0.8970   &  {\color{blue}32.32}/{\color{blue}0.9013}    &   32.27/0.9000  &  32.29/0.9010    &   {\color{red}32.34}/{\color{red}0.9015}   \\
			& $\times3$ &    27.21/0.7385          &    28.41/0.7863          &  28.83/0.7990    &       28.95/0.8004     &   28.96/0.8001 & 28.96/0.8010    &  {\color{blue}29.25}/{\color{red} 0.8093}    &    -/-   &  29.24/{\color{blue}0.8084}    &  {\color{red}29.29}/{\color{red}0.8093}     \\ 
			& $\times4$ &    25.96/0.6675          &     26.90/0.7101         &   27.29/0.7260   &    27.38/0.7284        &     27.40/0.7281 & 27.44/0.7313  &   27.71/{\color{blue} 0.7420}   &  {\color{blue} 27.72}/0.7400    &  {\color{blue} 27.72}/0.7409    &  {\color{red} 27.77}/{\color{red} 0.7419}      \\ \hline\hline
			\multirow{3}{*}{Urban100} & $\times2$ &  26.88/0.8403            &    29.50/0.8946          &  30.77/0.9140    &        31.23/0.9188    &    31.31/0.9195  & 31.41/0.9207  &  {\color{red}32.93}/{\color{red}0.9351}    &   32.55/0.9324   &  32.62/0.9328    &  {\color{blue}32.80}/{\color{blue}0.9341}     \\ 
			& $\times3$ &   24.46/0.7349           &     26.24/0.7989         &  27.14/0.8290    &       27.53/0.8378       &   27.56/0.8376 & 27.66/0.8415    &   {\color{blue}28.80}/{\color{blue} 0.8653}   &   -/-   &    28.73/0.8641  &   {\color{red} 28.89}/{\color{red} 0.8664}    \\
			& $\times4$ &   23.14/0.6577            &       24.52/0.7221       &    25.18/0.7540  &      25.44/0.7638       &    25.50//0.7630  & 25.71/0.7719  &   {\color{blue} 26.64}/{\color{blue} 0.8033}   &   26.38/0.7946   &  26.60/0.8015   & {\color{red} 26.73}/{\color{red} 0.8043}      \\ \hline\hline
			\multirow{3}{*}{Manga109} & $\times2$ &  30.30/0.9339            &     35.60/0.9663         &   37.22/0.9750   &    37.60/0.9736      &    37.72/0.9740  & 38.06/0.9757  &   {\color{blue}39.10}/0.9773   &   38.89/0.9775   &  39.08/{\color{blue} 0.9779}    &    {\color{red}39.28}/{\color{red}0.9784}  \\ 
			& $\times3$ &    26.95/0.8556          &      30.48/0.9117        &   32.01/0.9340   &      32.42/0.9359       &   32.51/0.9369  & 33.02/0.9404   &   34.17/0.9476   &  -/-    &   {\color{blue} 34.18}/{\color{blue} 0.9481}   &   {\color{red} 34.44}/{\color{red} 0.9494}    \\ 
			& $\times4$ &    24.89/0.7866          &       27.58/0.8555       &   28.83/0.8870   &      29.18/0.8914    &   29.42/0.8942 & 29.91/0.9008    &   31.02/0.9148   &  30.91/0.9137    &  {\color{blue} 31.15}/{\color{blue} 0.9160}   &   {\color{red} 31.40}/{\color{red} 0.9182}    \\ \hline
	\end{tabular}}
	\smallskip
	\caption{Average PSNR/SSIM values for scale factors $\times 2$, $\times 3$ and $\times 4$ with \textbf{BI} degradation model. The best performance is shown in {\color{red} red} and the second best performance is shown in {\color{blue} blue}.\label{comp_sot_bi}}
	
\end{table*}
	
	\begin{figure}[!htbp]
		\centering
		\includegraphics[width=.45\textwidth]{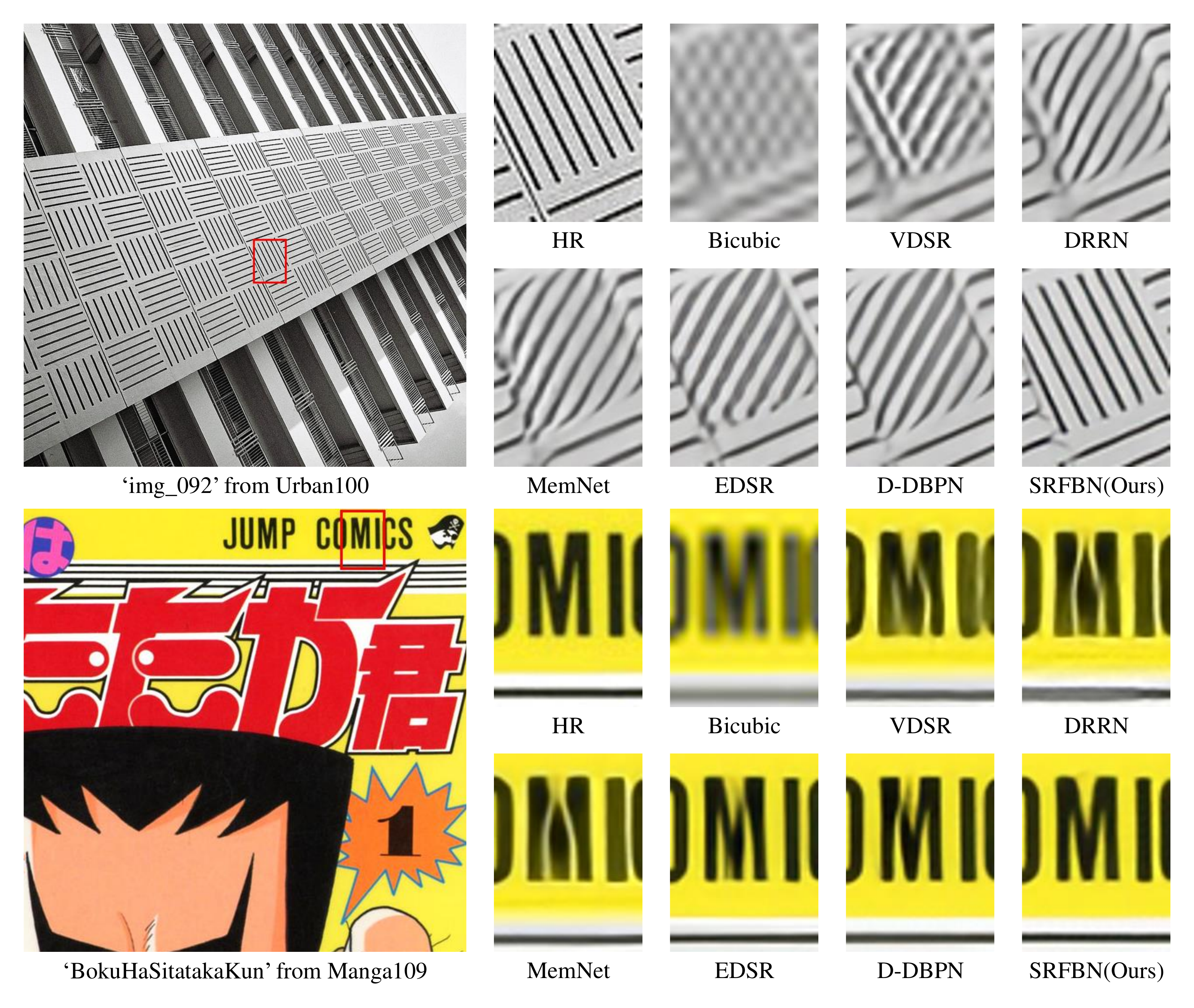}
		\caption{Visual results of \textbf{BI} degradation model with scale factor $\times 4$.}
		\label{visual_BI}
	\end{figure}

	\subsection{Comparison with the state-of-the-arts}
	The SRFBN with a larger base number of filters ($m$=64), which is derived from the SRFBN-L, is implemented for comparison. A self-ensemble method\cite{Timofte2016Seven} is also used to further improve the performance of the SRFBN (denoted as SRFBN+). A lightweight network SRFBN-S (T=4, G=3, $m$=32) is provided to compare with the state-of-the-art methods, which are carried only few parameters.
	
	\subsubsection{Network parameters}
	The state-of-the-art methods considered in this experiment include SRCNN\cite{dong2016image}, VDSR\cite{Kim_2016_CVPR}, DRRN\cite{Tai_2017_CVPR},  MemNet\cite{Tong_2017_ICCV}, EDSR\cite{lim2017enhanced}, DBPN-S\cite{Haris_2018_CVPR} and D-DBPN\cite{Haris_2018_CVPR}. The comparison results are given in Fig.~\ref{num_param} in terms of the network parameters and the reconstruction effects (PSNR). The SRFBN-S can achieve the best SR results among the networks with parameters fewer than 1000K. This demonstrates our method can well balance the number of parameters and the reconstruction performance. Meanwhile, in comparison with the networks with a large number of parameters, such as D-DBPN and EDSR, our proposed SRFBN and SRFBN+ can achieve competitive results, while only needs the 35\% and 7\% parameters of D-DBPN and EDSR, respectively. Thus, our network is lightweight and more efficient in comparison with other state-of-the-art methods. 
	
	\subsubsection{Results with BI degradation model}
	For \textbf{BI} degradation model, we compare the SRFBN and SRFBN+ with seven state-of-the-art image SR methods: SRCNN\cite{dong2016image}, VDSR\cite{Kim_2016_CVPR}, DRRN\cite{Tai_2017_CVPR}, SRDenseNet\cite{Tong_2017_ICCV}, MemNet\cite{Tong_2017_ICCV}, EDSR\cite{lim2017enhanced}, D-DBPN\cite{Haris_2018_CVPR}. The quantitative results in Tab.~\ref{comp_sot_bi} are re-evaluated from the corresponding public codes. Obviously, our proposed SRFBN can outperform almost all comparative methods. Compared with our method, EDSR utilizes much more number of filters (256 vs. 64), and D-DBPN employs more training images (DIV2K+Flickr2K+ImageNet vs. DIV2K+Flickr2K). However, our SRFBN can earn competitive results in contrast to them. In addition, it also can be seen that our SRFBN+ outperforms almost all comparative methods. 
	
	We show SR results with scale factor $\times 4$ in Fig.~\ref{visual_BI}. In general, the proposed SRFBN can yield more convincing results.  For the SR results of the `BokuHaSitatakaKun' image from Manga109, DRRN and MemNet even split the `M' letter. VDSR, EDSR and D-DBPN fail to recover the clear image. The proposed SRFBN produces a clear image which is very close to the ground truth. Besides, for the `img\_092' from Urban100, 
	the texture direction of the SR images from all comparative methods is wrong. However, our proposed SRFBN makes full use of the high-level information to take a self-correcting process, thus a more faithful SR image can be obtained. 
	
	\begin{figure}[!htbp]
	\centering
	\includegraphics[width=.45\textwidth]{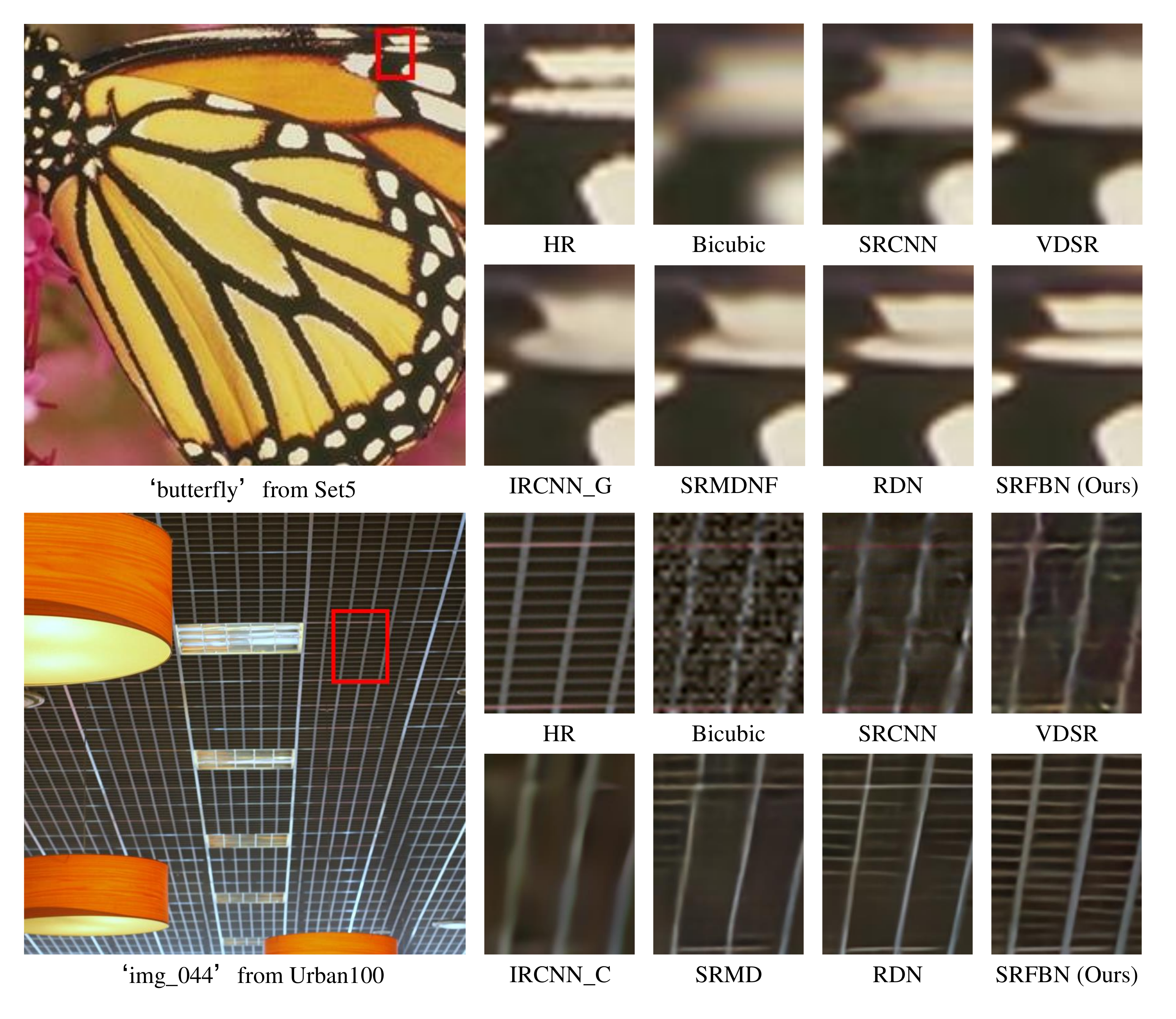}
	\caption{Visual results of \textbf{BD} and \textbf{DN} degradation models with scale factor $\times 3$. The first set of images shows the results obtained from \textbf{BD} degradation model. The second set of images shows the results from \textbf{DN} degradation model.}
	\label{visual_BD_DN}
	\vspace{-0.5cm}
	\end{figure}

	\begin{table*}[!htbp]	
	\centering	
	\resizebox{\textwidth}{!}{\begin{tabular}{|c|c|c|c|c|c|c|c|c|c|c|c|c|}
			\hline
			\multirow{2}{*}{Dataset} & \multirow{2}{*}{Model} & \multirow{2}{*}{Bicubic} & SRCNN & VDSR & IRCNN\_G                                   & IRCNN\_C                               & SRMD(NF)                                  & RDN                               & SRFBN                                      & SRFBN+  \\
			&                        &   &   \cite{dong2016image}                   &  \cite{Kim_2016_CVPR} &   \cite{DBLP:conf/cvpr/ZhangZGZ17}     & \cite{DBLP:conf/cvpr/ZhangZGZ17}   & \cite{Zhang2017Learning}  & \cite{Zhang_2018_CVPR}   & (Ours)  & (Ours) \\ \hline\hline
			\multirow{2}{*}{Set5}     & \textbf{BD}            & 28.34/0.8161             & 31.63/0.8888 & 33.30/0.9159 & 33.38/0.9182 & 29.55/0.8246 & 34.09/0.9242 & 34.57/0.9280                                      & \textcolor{blue}{34.66}/\textcolor{blue}{0.9283} & \textcolor{red}{34.77}/\textcolor{red}{0.9290}  \\
			& \textbf{DN}            & 24.14/0.5445             & 27.16/0.7672 & 27.72/0.7872 & 24.85/0.7205 & 26.18/0.7430 & 27.74/0.8026 & 28.46/0.8151                                      & \textcolor{blue}{28.53}/\textcolor{blue}{0.8182} & \textcolor{red}{28.59}/\textcolor{red}{0.8198}  \\ 
			\hline\hline
			\multirow{2}{*}{Set14}    & \textbf{BD}            & 26.12/0.7106             & 28.52/0.7924 & 29.67/0.8269 & 29.73/0.8292 & 27.33/0.7135 & 30.11/0.8364 & \textcolor{blue}{30.53}/\textcolor{blue}{0.8447 } & 30.48/0.8439                                     & \textcolor{red}{30.64}/\textcolor{red}{0.8458}  \\
			& \textbf{DN}            & 23.14/0.4828             & 25.49/0.6580 & 25.92/0.6786 & 23.84/0.6091 & 24.68/0.6300 & 26.13/0.6974 & \textcolor{blue}{26.60}/0.7101                    & \textcolor{blue}{26.60}/\textcolor{blue}{0.7144} & \textcolor{red}{26.67}/\textcolor{red}{0.7159}  \\ 
			\hline\hline
			\multirow{2}{*}{B100}     & \textbf{BD}            & 26.02/0.6733             & 27.76/0.7526 & 28.63/0.7903 & 28.65/0.7922 & 26.46/0.6572 & 28.98/0.8009 & \textcolor{blue}{29.23}/\textcolor{blue}{0.8079 } & 29.21/0.8069                                     & \textcolor{red}{29.28}/\textcolor{red}{0.8080}  \\
			& \textbf{DN}            & 22.94/0.4461             & 25.11/0.6151 & 25.52/0.6345 & 23.89/0.5688 & 24.52/0.5850 & 25.64/0.6495 & 25.93/0.6573                                      & \textcolor{blue}{25.95}/\textcolor{blue}{0.6625} & \textcolor{red}{25.99}/\textcolor{red}{0.6636}  \\ 
			\hline\hline
			\multirow{2}{*}{Urban100} & \textbf{BD}            & 23.20/0.6661             & 25.31/0.7612 & 26.75/0.8145 & 26.77/0.8154 & 24.89/0.7172 & 27.50/0.8370 & 28.46/\textcolor{blue}{0.8581}                    & \textcolor{blue}{28.48}/\textcolor{blue}{0.8581} & \textcolor{red}{28.68}/\textcolor{red}{0.8613}  \\
			& \textbf{DN}            & 21.63/0.4701             & 23.32/0.6500 & 23.83/0.6797 & 21.96/0.6018 & 22.63/0.6205 & 24.28/0.7092 & 24.92/0.7362                                      & \textcolor{blue}{24.99}/\textcolor{blue}{0.7424} & \textcolor{red}{25.10}/\textcolor{red}{0.7458}  \\ 
			\hline\hline
			\multirow{2}{*}{Manga109} & \textbf{BD}            & 25.03/0.7987             & 28.79/0.8851 & 31.66/0.9260 & 31.15/0.9245 & 28.68/0.8574 & 32.97/0.9391 & 33.97/0.9465                                      & \textcolor{blue}{34.07}/\textcolor{blue}{0.9466} & \textcolor{red}{34.43}/\textcolor{red}{0.9483}  \\
			& \textbf{DN}            & 23.08/0.5448             & 25.78/0.7889 & 26.41/0.8130 & 23.18/0.7466 & 24.74/0.7701 & 26.72/0.8424 & 28.00/0.8590                                      & \textcolor{blue}{28.02}/\textcolor{blue}{0.8618} & \textcolor{red}{28.17}/\textcolor{red}{0.8643}  \\
			\hline
			
	\end{tabular}}
	\smallskip
	\caption{Average PSNR/SSIM values for scale factor $\times3$ with \textbf{BD} and \textbf{DN} degradation models. The best performance is shown in {\color{red}red} and the second best performance is shown in {\color{blue}blue}.\label{comp_sot_md}}
	
	\end{table*}
	\subsubsection{Results with BD and DN degradation models}
	\label{4.5.3}
	As aforementioned, the proposed SRFBN is trained using curriculum learning strategy for \textbf{BD} and \textbf{DN} degradation models, and fine-tuned based on \textbf{BI} degradation model using DIV2K. The proposed SRFBN and SRFBN+ are compared with SRCNN\cite{dong2016image}, VDSR\cite{Kim_2016_CVPR}, IRCNN\_G\cite{DBLP:conf/cvpr/ZhangZGZ17}, IRCNN\_C\cite{DBLP:conf/cvpr/ZhangZGZ17}, SRMD(NF)\cite{Zhang2017Learning}, and RDN\cite{Zhang_2018_CVPR}. Because of degradation mismatch, SRCNN and VDSR are re-trained for \textbf{BD} and \textbf{DN} degradation models. As shown in Tab.~\ref{comp_sot_md}, The proposed SRFBN and SRFBN+ achieve the best on almost all quantative results over other state-of-the-art methods.
	
	In Fig.~\ref{visual_BD_DN}, we also show two sets of visual results with \textbf{BD} and \textbf{DN} degradation models from the standard benchmark datasets. Compared with other methods, the proposed SRFBN could alleviate the distortions and generate more accurate details in SR images. From above comparisions, we further indicate the robustness and effectiveness of SRFBN in handling \textbf{BD} and \textbf{DN} degradation models. 
	
	\section{Conclusion}
	In this paper, we propose a novel network for image SR called super-resolution feedback network (SRFBN) to faithfully reconstruct a SR image by enhancing low-level representations with high-level ones. The feedback block (FB) in the network can effectively handle the feedback information flow as well as the feature reuse. In addition, a curriculum learning strategy is proposed to enable the network to well suitable for more complicated tasks, where the low-resolution images are corrupted by complex degradation models. The comprehensive experimental results have demonstrated that the proposed SRFBN could deliver the comparative or better performance in comparison with the state-of-the-art methods by using very fewer parameters. 
	\newline
	
	\noindent\textbf{Acknowledgement}. The research in our paper is sponsored by National Natural Science Foundation of China (No.61701327 and No.61711540303), Science Foundation of Sichuan Science and Technology Department (No.2018GZ0178).
	\clearpage
	{\small
		\bibliographystyle{ieee}
		\bibliography{ms}
	}

\clearpage
\appendix
\noindent\textbf{\large{Supplementary Material}} 

\vspace{0.3cm}
\textit{The following items are contained in the supplementary material:}

\textit{1. Discussions on the feedback block.}

\textit{2. More insights on the feedback mechanism.}

\textit{3. Quantitative results using DIV2K training images.}

\textit{4. Running time comparison.}

\textit{5. More qualitative results.}

\section{Study of Feedback Block}
More effective basic block could generate finer high-level representations and then benefit our feedback process. Thus, we explore the design of the basic block in this section. We still use SRFBN-L (T=4, G=6), which has a small base number of fiters ($m$=32) for analysis. 

\textbf{Ablation study} mainly focuses on two components of our feedback block (FB): (1) up- and down-sampling layers (UDSL), (2) dense skip connecitons (DSC). To analysis the effect of UDSL in our proposed FB, we replace the up- and down-sampling layers with $3\times3$ sized convolutional layers (with one padding and one stridding). In Tab.~\ref{ab_block}, when UDSL is replaced with $3\times3$ sized convolutional layers in the FB, the PSNR value dramatically decreases. This indicates that up- and down-sampling operations carrying large kernel size can expliot abundant contextual information and are effective for image super-resolution (SR). After adding DSC to the FB, the reconstruction performance can be further improved, because the information efficiently flows through DSC across hierarchy layers and even across time.
\begin{table}[h]
	\begin{center}
		\begin{tabular}{|c|c|c|c|c|}
			\hline
			\multicolumn{5}{|c|}{Different combinations of UDSL and DSC}                                                                   \\ \hline \hline
				UDSL &  \ding{55}    &   \ding{52}        & \ding{55}        & \ding{52}     \\ \hline
				DSC &  \ding{55}    &   \ding{55}        &  \ding{52}       &  \ding{52}     \\ \hline \hline
				PSNR & 31.41	 &    32.05        &  31.62	      &   \textbf{32.11}     \\ \hline
		\end{tabular}
		\medskip
		\caption{The investigation of up- and down-sampling layers (UDSL), and dense skip connection (DSC) with scale factor $\times4$ on Set5.}
		\label{ab_block}
		\vspace{-0.5cm}			
	\end{center}
\end{table}

\textbf{Other basic blocks} are considered in this experiment in comparison with our FB. We choose two superior basic blocks (\ie projection units\cite{Haris_2018_CVPR} and RDB\cite{Zhang_2018_CVPR}), which were designed for image SR task recently, and ConvLSTM from \cite{Zamir_2017_CVPR} for comparison. To keep consistency, the number of convolutional layers\footnote{$1\times1$ sized convolutional layers are omitted.} and filters in each basic block are set to 12 and 32, respectively. In Tab.~\ref{block_design}, we first see that all SR custom basic blocks outperform ConvLSTM by a large margin. The gate mechanisms in ConvLSTM influence the distribution and intensity of original images and thus are difficult to meet high fidelity needs in image SR tasks. Besides, high-level information is directly added to low-level information in ConvLSTM, causing the loss of enough contextual information for the next iteration. Noticeably, our proposed FB obtains the best quantitative results in comparison with other basic blocks. This further demonstrates the powerful representation ability of our proposed FB. 

\begin{table*}[h]
	\begin{center}
		\begin{tabular}{|c|c|c|c|c|}
			\hline
    		 & ConvLSTM & Projection units  & RDB & Ours \\ \hline\hline
		PSNR &    31.26	      &    32.07   	       &   32.01	       &  \textbf{32.11}    \\ \hline
		\end{tabular}
		\medskip
		\caption{The investigation of other block design with scale factor $\times4$ on Set5.}
		\label{block_design}			
	\end{center}
\end{table*}

\begin{table*}[h]
	\begin{center}
		\begin{tabular}{|c|c|c|c|c|c|c|}
			\hline
			& Params. & Set5        & Set14       & B100        & Urban100    & Manga109    \\ \hline\hline
			MemNet-Pytorch  &  677K &  31.75/0.889           &   28.31/0.775          &  27.37/0.729           &   25.54/0.766          &  29.65/0.897           \\ \hline
			D-DBPN \cite{Haris_2018_CVPR} &  10,426K   & \textbf{32.40/0.897} & 28.75/0.785 & 27.67/0.738 & 26.38/0.793 & 30.89/0.913 \\ \hline
			SRFBN-S (Ours) &  483K &  31.98/0.892           &   28.45/0.778          &  27.44/0.731          &   25.71/0.772          &  29.91/0.901         \\ \hline
			SRFBN (Ours) &  3,631K   &   32.39/\textbf{0.897}        &    \textbf{28.77}/\textbf{0.786}         &    \textbf{27.68}/\textbf{0.740}         &       \textbf{26.47}/\textbf{0.798 }    &  \textbf{30.96}/\textbf{0.914} \\ \hline         
		
		\end{tabular}
		\medskip
		\caption{Average PSNR/SSIM values for scaling factor $\times4$ using \textbf{BI} degradation model. The networks used for comparison are all trained using DIV2K training images. The best performance is \textbf{highlighted}.}
		\label{div2k_comp}			
	\end{center}
\end{table*}

\section{Additional Insights on Feedback Mechanism}
\begin{figure}[t]
	\centering
	\includegraphics[width=.45\textwidth]{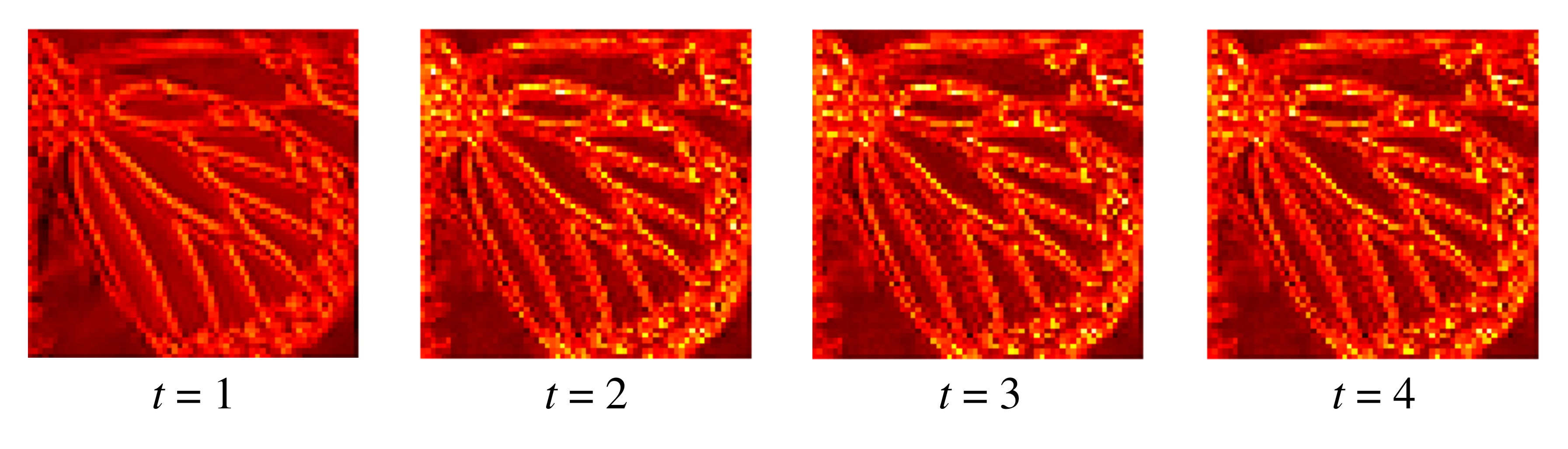}
	\caption{Average feature maps of refined low-level features from different iterations in the propose SRFBN (zoom for a better view). All average feature maps use the same colormap for better visualization.}
	\label{L0_vis}
\end{figure}

\begin{figure}[t]
	\centering
	\includegraphics[width=.48\textwidth]{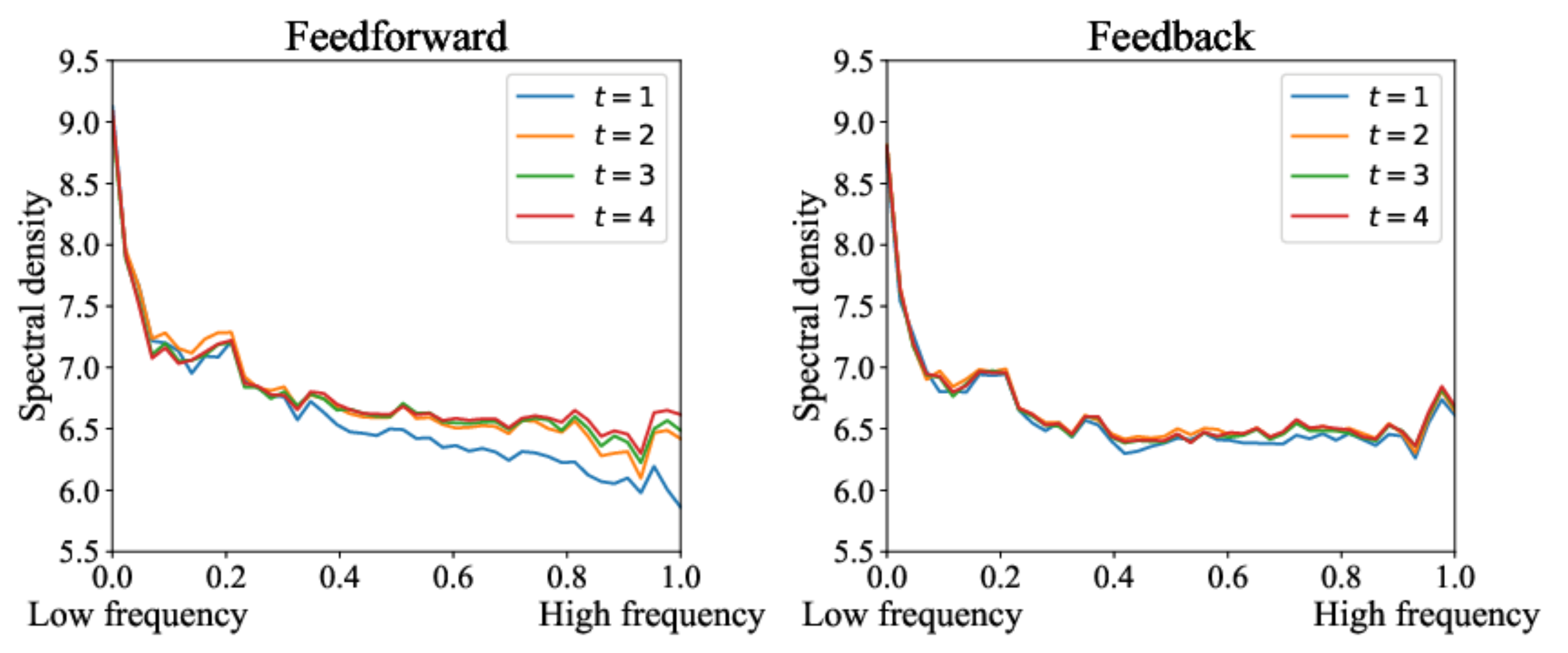}
	\caption{The spectral densities of the average feature map at each iteration $t$ (zoom for a better view). From left to right of the horizontal axis, frequency is normalized and ranged from low to high for better visualization.}
	\label{spec_comp}
\end{figure}

For better understanding the feedback mechanism in the proposed network, we visualize the average feature map of $L_{0}^{t}$ at each iteration $t$ in Fig.~\ref{L0_vis}. $L_{0}^{t}$ actually represents the low-level representations refined by high-level features $F_{out}^{t-1}$ from last iteration (see the main paper's Eq. 5). The initial state $F_{out}^{0}$ is set to $F_{in}^{1}$, hence the first iteration in the proposed network can not receive the feedback information. From Fig.~\ref{L0_vis}, we observe that, except the first iteration ($t$=1), these average feature maps show bright activations in the contours and outline edges of the original image. It seems that the feedback connection adds the high-level representations to the initial feature maps. This further indicates that initial low-level features, which lack enough contextual information, surely are corrected using high-level information through the feedback mechanism in the proposed network. 

To further conduct differences between feedforward and feedback networks, we plot 1-D spectral densities of the average feature map at each iteration $t$ in SRFBN-L (feedback) and SRFBN-L-FF (feedforward). As shown in Fig.5 of the main paper, each average feature map is the mean of $F_{out}^{t}$. To acquire 1-D spectral densities of the average feature map at each iteration $t$, we get the 2-D spectrum map through discrete Fourier transform, center the low-frequency component of the spectrum map, and place concentric annular regions to compute the mean of spectral densities for continuous frequency ranges. From Fig.~\ref{spec_comp}, we can conclude that the feedback network can estimate more mid-frequency and high-frequency information than the feedforward network at early iterations. With the iteration $t$ grows, the feedforward network gradually recovers mid-frequency and high-frequency components, while the feedback network pays attention to refine the well-developed information. For the feedback network, we also observe that,  because of the help of the feedback mechanism ($t>$1), mid-frequency and high-frequency information of the average feature map at the second iteration ($t$=2) is more similiar to the final representations ($t$=4) in contrast to the first iteration ($t$=1).

\section{Sanity Check}
To purely investigate the effect of the network architecture design, we compare the quantitative results obtained from different networks using the same training dataset (DIV2K training images\cite{Agustsson_2017_CVPR_Workshops}). The choices of networks for comparison include D-DBPN (which is a state-of-the-art network with moderate parameters) and MemNet\cite{tai2017memnet} (which is the leading network with recurrent structure). Because MemNet only reveals the results trained using 291 images, we re-train it using DIV2K on Pytorch framework. The results of D-DBPN are cited from their supplementary materials. Our SRFBN-S (T=4, G=3, $m$=32) and final SRFBN (T=4, G=6, $m$=64) are provided for this comparison. In Tab.~\ref{div2k_comp}, our SRFBN-S shows better quantitative results than MemNet with 71\% fewer parameters. Moreover, the final SRFBN also gains competitive results in contrast to D-DBPN especially on Urban100 and Manga109 datasets, which mainly contain images with a large size. This comparison shows the effectiveness of the proposed SRFBN.

\begin{table}[htbp]
	\centering
	\begin{tabular}{|c|c|c|}
		\hline
		Model          & Running time (s) & PSNR  \\ \hline \hline
		MemNet-Pytorch & 0.481            & 25.54 \\ \hline
		EDSR           & 1.218            & 26.64 \\ \hline
		D-DBPN         & 0.015            & 26.38 \\ \hline
		RDN            & 1.268            & 26.61 \\ \hline
		RCAN           & 1.130            & 26.82 \\ \hline
		SRFBN-S (Ours) & 0.006            & 25.71 \\ \hline
		SRFBN (Ours)   & 0.011            & 26.60 \\ \hline
	\end{tabular}
	\medskip
	\caption{Average running time comparison on Urban100 with scale factor 4 on an NVIDIA 1080Ti GPU.}
	\label{runtime}
\end{table}

\section{Running Time Comparison}
 We compare running time of our proposed SRFBN-S and SRFBN with five state-of-the-art networks: MemNet\cite{tai2017memnet}, EDSR\cite{lim2017enhanced}, D-DBPN\cite{Haris_2018_CVPR}, RDN\cite{Zhang_2018_CVPR} and RCAN\cite{zhang2018rcan} on Urban100 with scale factor $\times4$. Because the large memory consumption in Caffe, we re-implement MemNet on Pytorch for fair comparison. The running time of all networks is evaluated on the same machine with 4.2GHz Intel i7 CPU (16G RAM) and an NVIDIA 1080Ti GPU using their official codes. Tab.~\ref{runtime} shows that our SRFBN-S and SRFBN have the fastest evaluation time in comparison with other networks. This further reflects the effectiveness of our proposed networks. The quantitative results of our proposed networks are less comparable with RCAN, but RCAN mainly focuses on much deeper networks design (about 400 convolutional layers) to purchase more accurate SR results. In contrast, our SRFBN only has about 100 convolutional layers with 77\% fewer parameters (3,631K vs. 15,592K) than RCAN.

\section{More Qualitative Results}
\begin{figure*}[h]
	\centering
	\includegraphics[width=.9\textwidth]{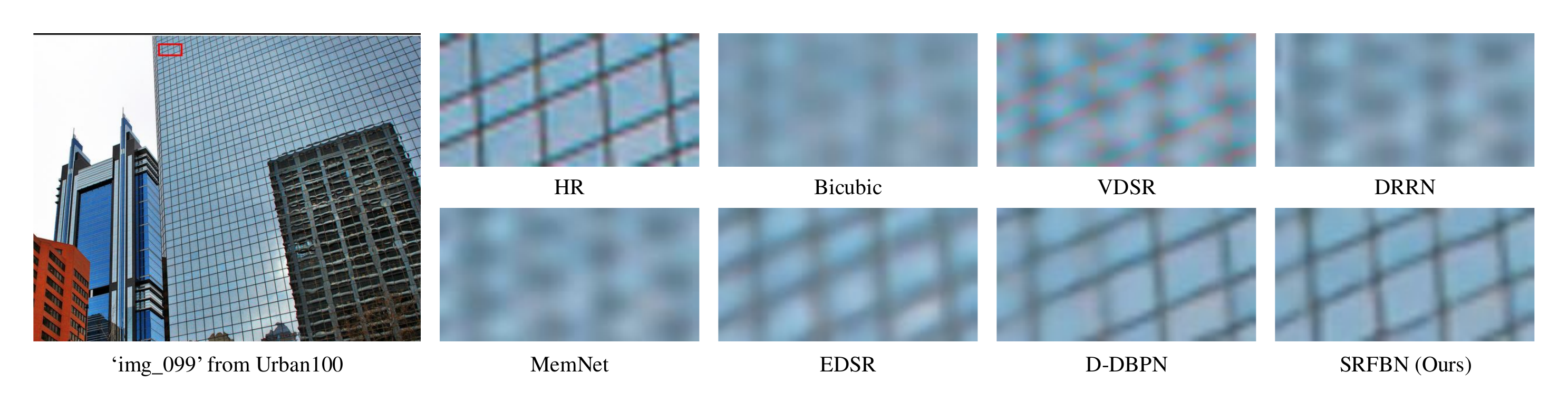}
	\caption{Visual results of \textbf{BI} degradation model with scale factor $\times 4$.}
	\label{figure2}
\end{figure*}

In Fig.~\ref{figure2}-\ref{figure13}, we provide more visual results of different degradation models to prove the superiority of the proposed network.


\begin{figure*}[h]
	\centering
	\includegraphics[width=.9\textwidth]{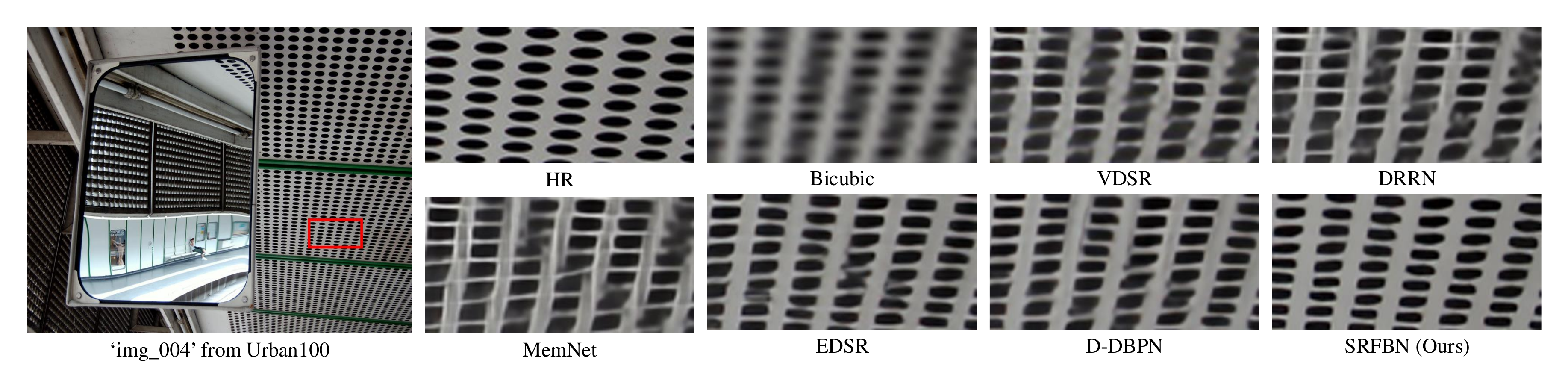}
	\caption{Visual results of \textbf{BI} degradation model with scale factor $\times 4$.}
\end{figure*}
\begin{figure*}[h]
	\centering
	\includegraphics[width=.9\textwidth]{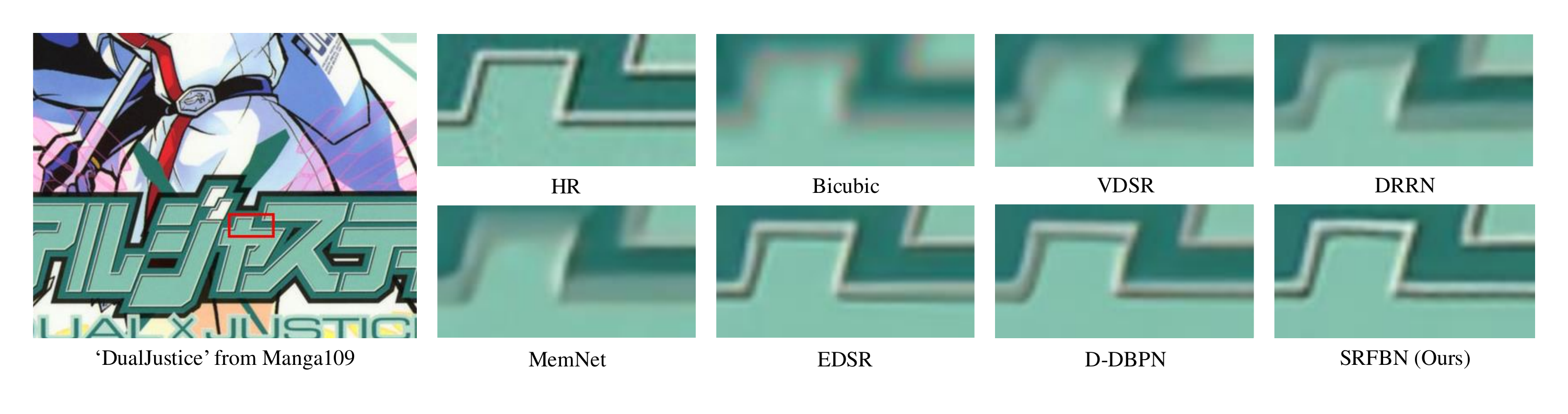}
	\caption{Visual results of \textbf{BI} degradation model with scale factor $\times 4$.}
\end{figure*}
\begin{figure*}[h]
	\centering
	\includegraphics[width=.9\textwidth]{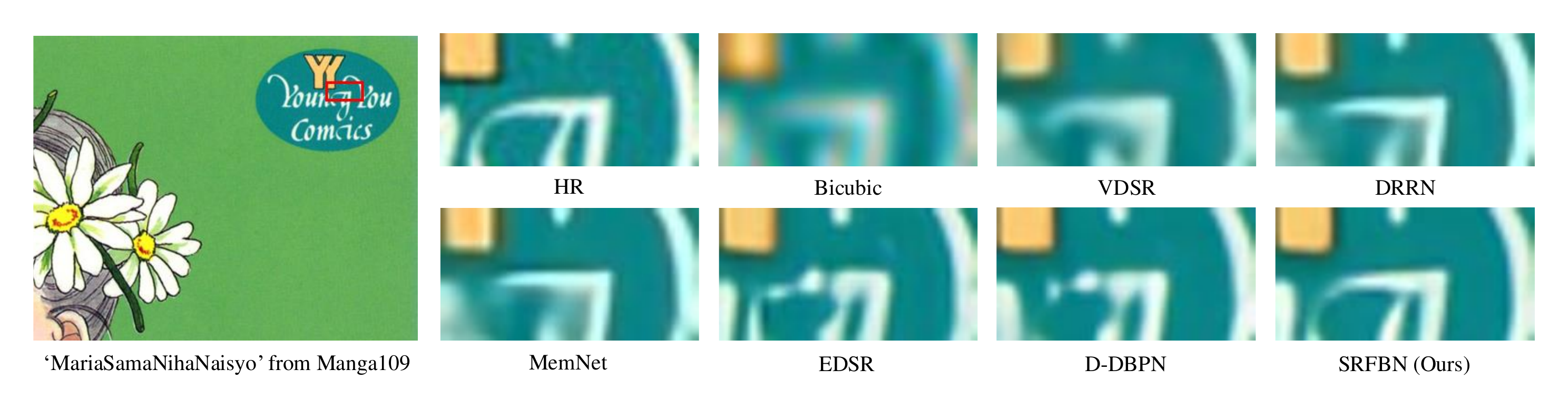}
	\caption{Visual results of \textbf{BI} degradation model with scale factor $\times 4$.}
\end{figure*}


\begin{figure*}[h]
	\centering
	\includegraphics[width=.9\textwidth]{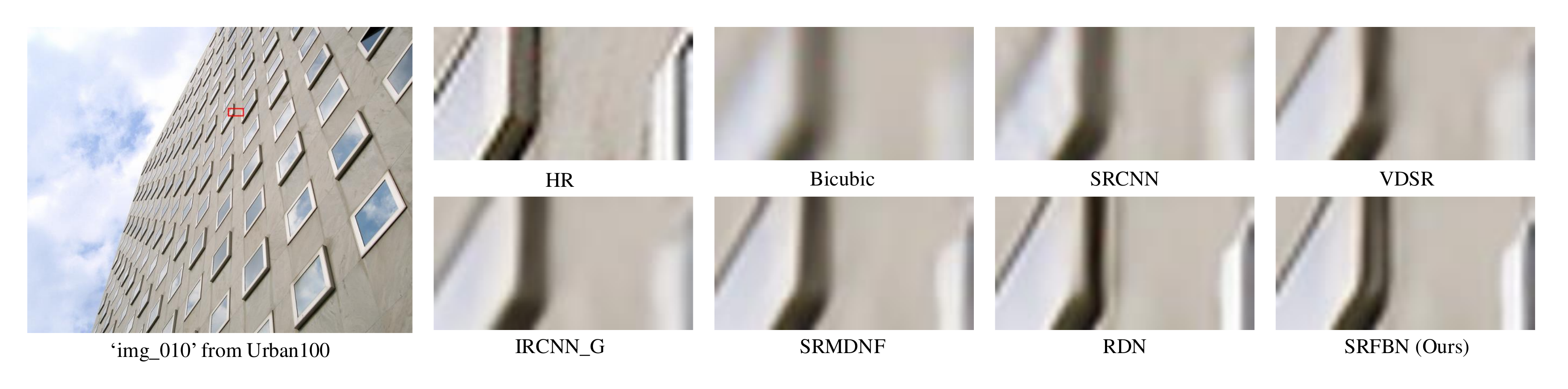}
	\caption{Visual results of \textbf{BD} degradation model with scale factor $\times 4$.}
\end{figure*}
\begin{figure*}[h]
	\centering
	\includegraphics[width=.9\textwidth]{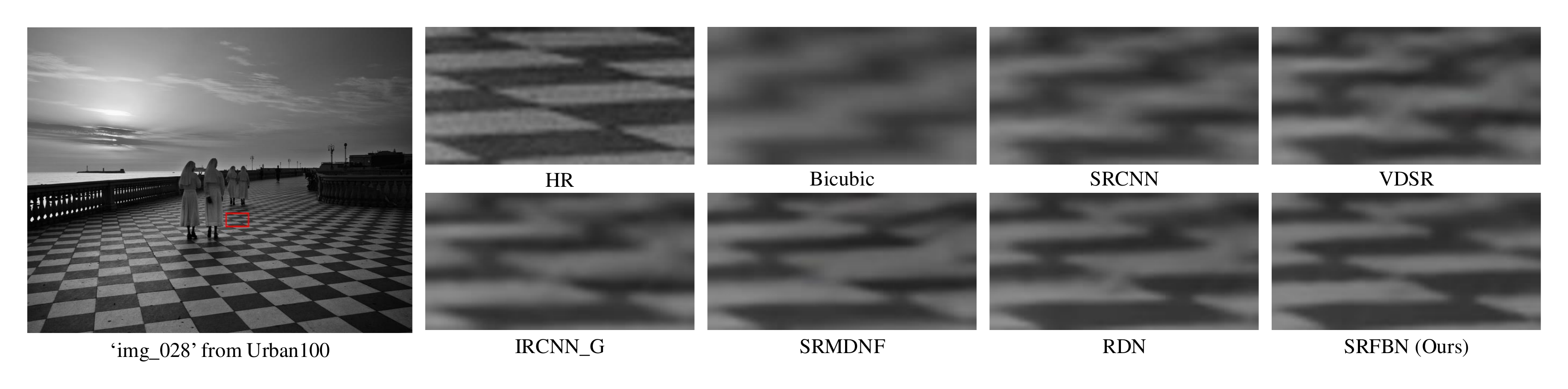}
	\caption{Visual results of \textbf{BD} degradation model with scale factor $\times 4$.}
\end{figure*}
\begin{figure*}[h]
	\centering
	\includegraphics[width=.9\textwidth]{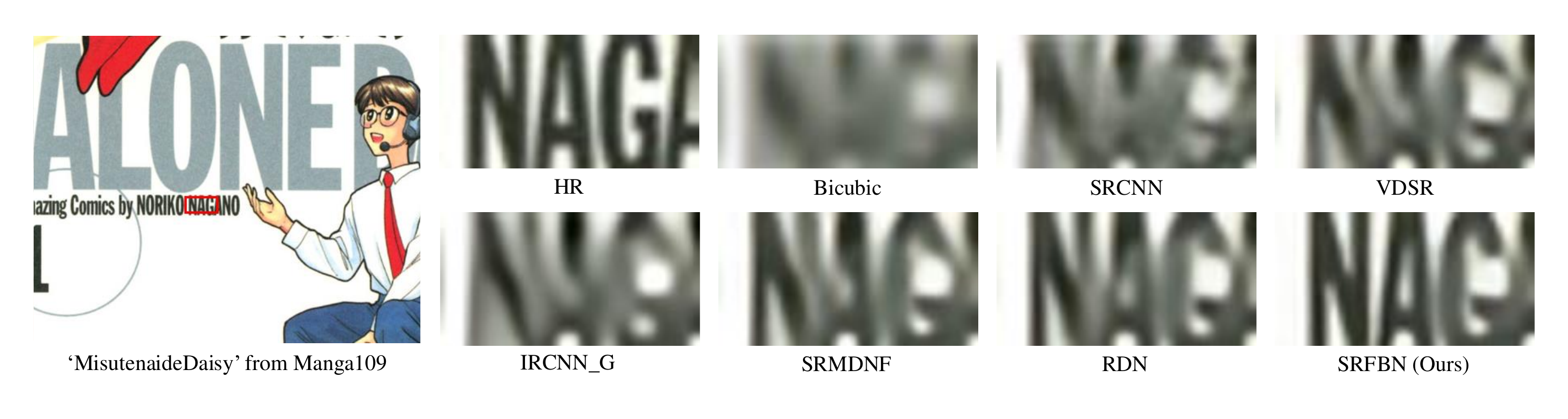}
	\caption{Visual results of \textbf{BD} degradation model with scale factor $\times 4$.}
\end{figure*}
\begin{figure*}[h]
	\centering
	\includegraphics[width=.9\textwidth]{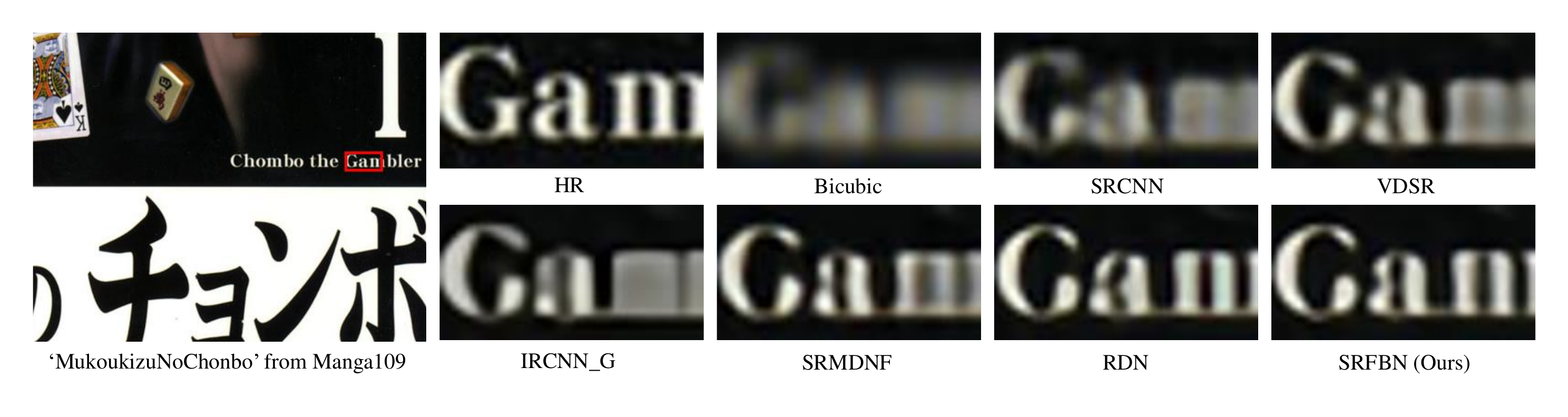}
	\caption{Visual results of \textbf{BD} degradation model with scale factor $\times 4$.}
\end{figure*}


\begin{figure*}[h]
	\centering
	\includegraphics[width=.9\textwidth]{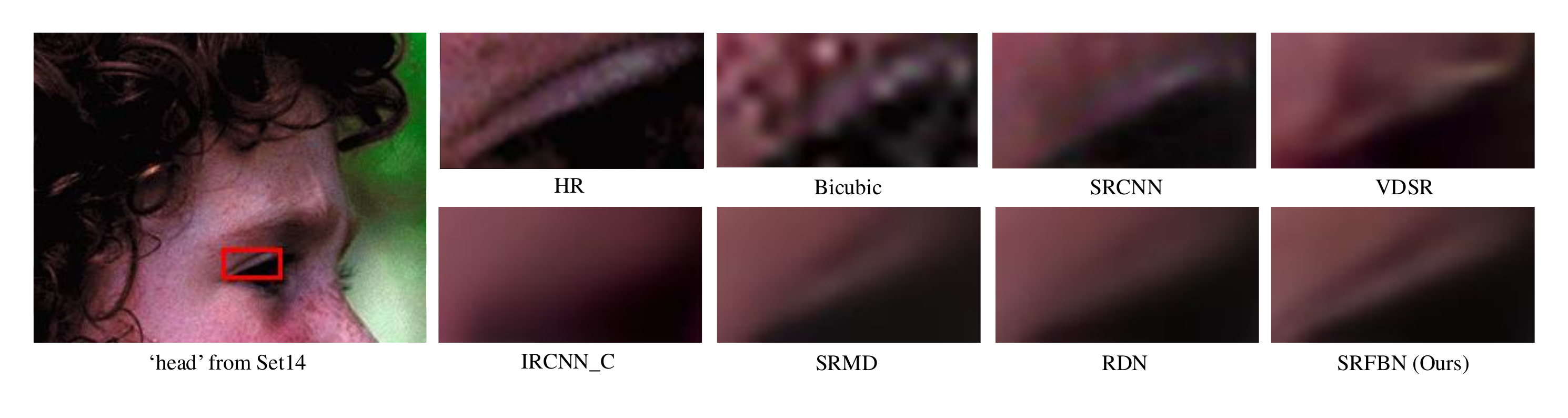}
	\caption{Visual results of \textbf{DN} degradation model with scale factor $\times 4$.}
\end{figure*}
\begin{figure*}[h]
	\centering
	\includegraphics[width=.9\textwidth]{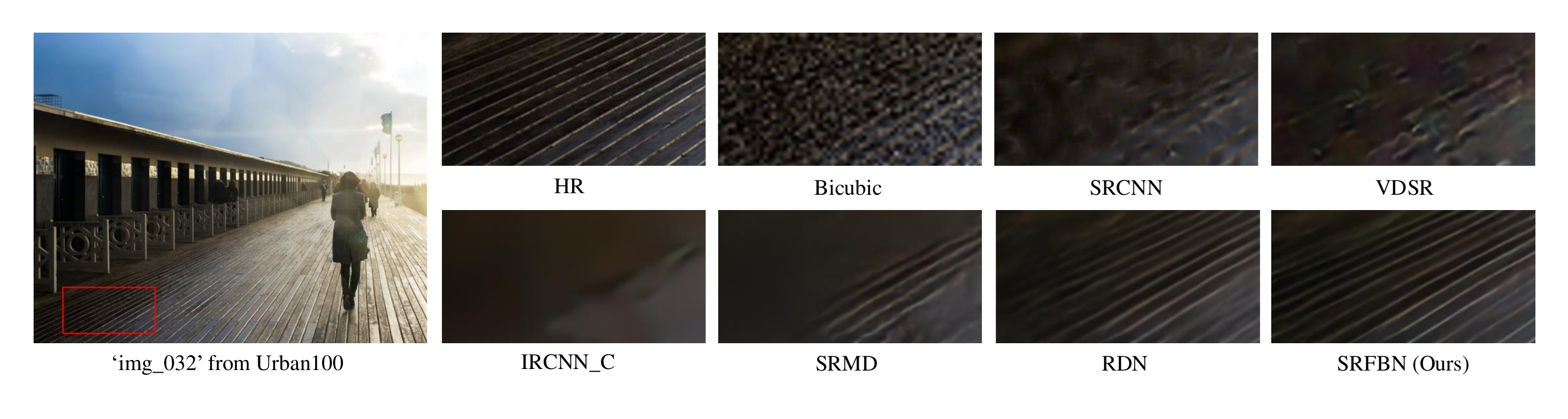}
	\caption{Visual results of \textbf{DN} degradation model with scale factor $\times 4$.}
\end{figure*}
\begin{figure*}[h]
	\centering
	\includegraphics[width=.9\textwidth]{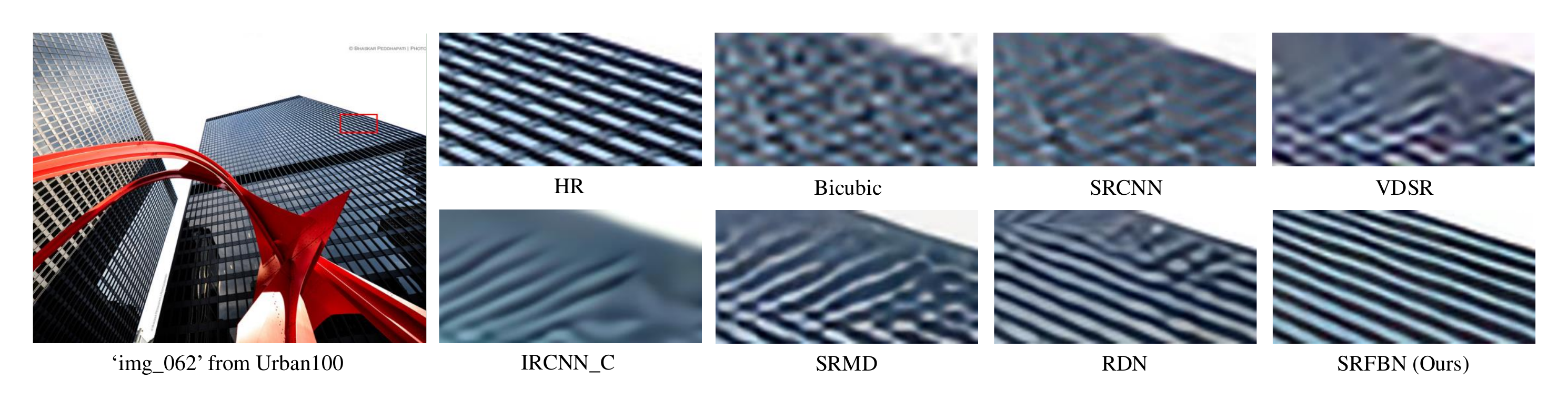}
	\caption{Visual results of \textbf{DN} degradation model with scale factor $\times 4$.}
\end{figure*}
\begin{figure*}[h]
	\centering
	\includegraphics[width=.9\textwidth]{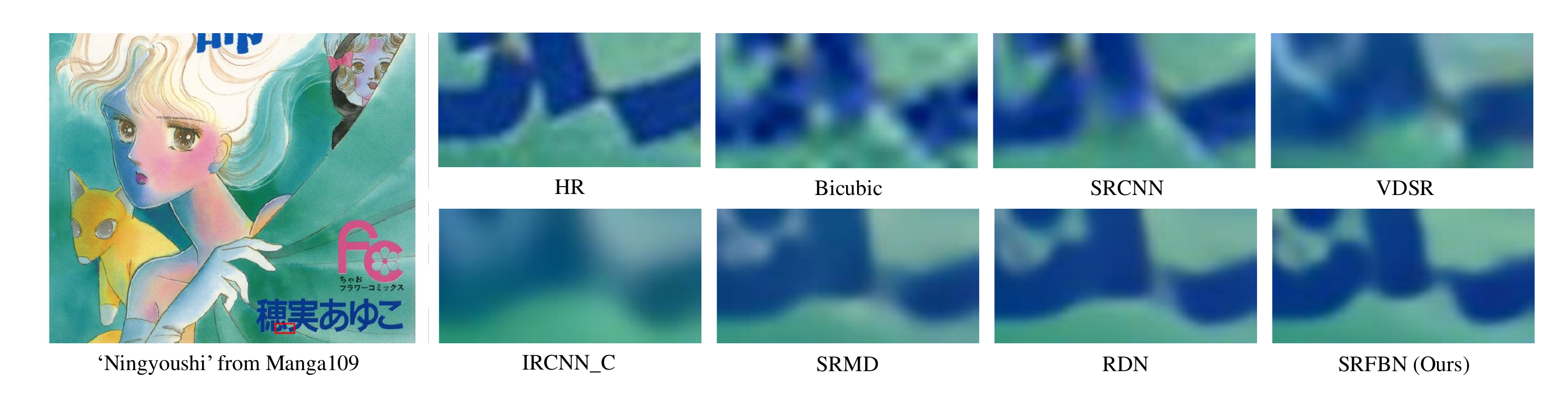}
	\caption{Visual results of \textbf{DN} degradation model with scale factor $\times 4$.}
	\label{figure13}	
\end{figure*}

\end{document}